\documentclass[10pt,journal,compsoc]{IEEEtran}
\usepackage{float}
\usepackage{graphicx} % for pdf, bitmapped graphics files
\usepackage{epsfig} % for postscript graphics files
\usepackage{times} % assumes new font selection scheme installed
\usepackage{amsmath} % assumes amsmath package installed
\usepackage{amssymb}  % assumes amsmath package installed
\usepackage{amsfonts,amssymb,amsmath,latexsym}
\usepackage{multirow,multicol}
\usepackage[ruled,linesnumbered]{algorithm2e}
\usepackage{algpseudocode}
\usepackage{hyperref}%can find the location of current ref. 
\usepackage{epstopdf} %transfer eps to pdf
\usepackage{subfigure} %for subfigures
\usepackage{stfloats} %for table
\usepackage{bm}
\usepackage{color}
\usepackage{amsthm}
\usepackage{array}
\usepackage{threeparttable}

\usepackage{url}  
\newtheorem{theorem}{\indent Theorem}

\theoremstyle{remark}
\newtheorem{remark}{\indent Remark}
\ifCLASSOPTIONcompsoc
  \usepackage[nocompress]{cite}
\else
  \usepackage{cite}
\fi
\ifCLASSINFOpdf
\else
\fi
\begin{document}

\title{MIXRTs: Toward Interpretable Multi-Agent Reinforcement Learning via Mixing Recurrent Soft Decision Trees}

\author{Zichuan Liu,~Yuanyang~Zhu,~\IEEEmembership{Member,~IEEE,}~Zhi~Wang,~\IEEEmembership{Member,~IEEE,}~Yang~Gao,~\IEEEmembership{Senior Member,~IEEE,} Chunlin~Chen,~\IEEEmembership{Senior Member,~IEEE}
	
\IEEEcompsocitemizethanks{
\IEEEcompsocthanksitem Z. Liu, Z. Wang and C. Chen are with the Department of Control Science and Intelligent Engineering, School of Management and Engineering, Nanjing University, Nanjing 210093, China. E-mail: zichuanliu@smail.nju.edu.cn, \{zhiwang, clchen\}@nju.edu.cn.\protect\\

\IEEEcompsocthanksitem Y. Zhu is with the Laboratory
of Data Intelligence and Interdisciplinary Innovation, School of Information Management, Nanjing University, Nanjing 210023, China. E-mail: yuanyangzhu@nju.edu.cn. \protect\\

\IEEEcompsocthanksitem Y. Gao is with the State Key Laboratory for Novel Software Technology, Nanjing University, Nanjing 210023, China. E-mail: gaoy@nju.edu.cn.}

\thanks{This work was supported in part by the National Natural Science Foundation of China under Grant 62376122, Grant 62073160, and Grant 72394363, 
in part by the Nanjing University Integrated Research Platform of the Ministry of Education-Top Talents Program,
in part by the AI \& AI for Science Project of Nanjing University, 
and in part by the Jiangsu Science and Technology Major Project BG2024031. (Corresponding authors: Yuanyang Zhu and Zhi Wang.)}
}

\markboth{IEEE Transactions on Pattern Analysis and Machine Intelligence}
{Shell \MakeLowercase{\textit{et al.}}: Bare Demo of IEEEtran.cls for Computer Society Journals}

\IEEEtitleabstractindextext{%
\begin{abstract}
While achieving tremendous success in various fields, existing multi-agent reinforcement learning (MARL) with a black-box neural network makes decisions in an opaque manner that hinders humans from understanding the learned knowledge and how input observations influence decisions. In contrast, existing interpretable approaches usually suffer from weak expressivity and low performance. To bridge this gap, we propose MIXing Recurrent soft decision Trees (MIXRTs), a novel interpretable architecture that can represent explicit decision processes via the root-to-leaf path and reflect each agent's contribution to the team. Specifically, we construct a novel soft decision tree using a recurrent structure and demonstrate which features influence the decision-making process. Then, based on the value decomposition framework, we linearly assign credit to each agent by explicitly mixing individual action values to estimate the joint action value using only local observations, providing new insights into interpreting the cooperation mechanism. Theoretical analysis confirms that MIXRTs guarantee additivity and monotonicity in the factorization of joint action values. Evaluations on complex tasks like Spread and StarCraft II demonstrate that MIXRTs compete with existing methods while providing clear explanations, paving the way for interpretable and high-performing MARL systems.
\end{abstract}

}

\maketitle

\IEEEdisplaynontitleabstractindextext

\IEEEpeerreviewmaketitle

\IEEEraisesectionheading{\section{Introduction}\label{sec:introduction}}

\IEEEPARstart{M}{ulti-agent} reinforcement learning (MARL) has been shown considerable potential in solving a variety of challenging tasks, e.g., games~\cite{vinyals2019grandmaster,sunehag2017value,rashid2018qmix}, autonomous driving~\cite{li2022metadrive,yu2019distributed,kiran2021deep} and robotics interactions~\cite{liu2021semantic,10250993}. 
Despite these promising results, much of this progress relies on deep neural network (DNN) models serving as powerful function approximators, encoded with thousands to millions of parameters interacting in complex and nonlinear ways~\cite{topin2021iterative}.
This architectural complexity brings substantial obstacles for humans to understand how decisions are made and what key features influence decisions, especially when the network becomes deeper in size or more complex structures are appended~\cite{liu2019tabby,zhou2018interpreting}.
Indeed, creating mechanisms to interpret the implicit behaviors of black-box DNNs remains an open problem in the field of machine learning~\cite{hassija2024interpreting, tjoa2020survey}.

It is crucial to gain insights into the decision-making process of artificially intelligent agents for their successful and reliable deployment into real-world applications, especially in high-risk domains such as healthcare and military~\cite{rudin2019stop, rudin2022interpretable, xu2022towards, cao2019interpretable}.
The lack of transparency in MARL techniques imposes significant limitations on practitioners, bringing critical barriers to establishing trust in the learned policies and scrutinizing their weaknesses~\cite{natarajan2020effects}.
Explainable reinforcement learning~\cite{puiutta2020explainable,shi2020self, zheng2024symbolic } emerges as a promising approach to develop transparent procedures that can be followed step-by-step or held accountable by human operators.
While existing explainable methods offer some potential~\cite{verma2019verifiable, jiang2019neural} or vision-based interpretations~\cite{shi2021temporal, zheng2024symbolic} in single-agent tasks, they still struggle to balance interpretability and performance in complex reinforcement learning tasks~\cite{silver2016mastering}, especially in multi-agent domains.

Traditional decision trees~\cite{loh2011classification,breiman2001random} provide interpretable inferences at a rule level, as humans can easily understand their decision process by visualizing decision paths.
However, they often suffer from limited expressivity and low accuracy when using shallow trees and univariate decision nodes, leading to a difficult trade-off between model interpretability and performance.
Alternatively, differentiable soft decision trees (SDTs)~\cite{frosst2017distilling, silva2020optimization} are built on the structure of fuzzy decision trees, bridging traditional decision trees and neural networks in terms of expressiveness~\cite{suarez1999globally}.
SDTs offer interpretations that non-expert individuals can easily visualize and simulate, thereby enhancing human readability.
Several works~\cite{coppens2019distilling, ding2020cdt} have attempted to use an imitation learning paradigm to distill a pre-trained DNN control policy into an SDT, providing an interpretable form of policy in single-agent tasks.
However, the simple structure of an SDT often makes it difficult to accurately mimic its original policy.
Therefore, it remains a challenge to find a ``sweet spot'' that can well balance fidelity and simplicity in tree models.

Rather than imitating a pre-trained policy, an alternative paradigm is to train an SDT policy from agent experience in an end-to-end manner, which directly learns the domain knowledge from tasks using interpretable models. 
While SDT has achieved an adequate balance between interpretability and performance in simple single-agent domains (e.g., CartPole and MountainCar in OpenAI Gym~\cite{brockman2016openai})~\cite{ding2020cdt}, they cannot maintain satisfying learning performance without sacrificing interpretability in complex multi-agent tasks due to limited model expressivity.
Particularly, to the best of our knowledge, there is seldom existing work exploring the model interpretability in MARL domains.
Here, our goal is to strike a favorable balance between interpretability and insight into the underlying decision-making process using tree-based models in a multi-agent system.

In this paper, we propose a novel MIXing Recurrent soft decision Trees (MIXRTs) method to tackle the tension between model interpretability and learning performance in MARL domains.
Instead of attempting to understand how a DNN makes its decisions, we utilize a differentiable SDT to learn the decision-making process for a certain task.
First, to facilitate learning over long timescales for each agent, we propose the recurrent tree cell (RTC) that receives the current individual observation and history information as input at each timestamp.
Second, we utilize a linear combination of multiple RTCs to improve performance and reduce the variance of SDTs, while maintaining the interpretability of RTCs.
By visualizing the tree structure, RTCs can provide intuitive explanations of how important features affect the decision process.
MIXRTs consist of RTCs representing the individual value function and a mixing tree structure aiming to learn an optimal linear value decomposition, which ensures consistency between the centralized and decentralized policies.
The linear mixing structure emphasizes the explanations about what role each agent plays in cooperative tasks by analyzing its assigned credit.
To improve learning efficiency, we also use parameter sharing across individual RTCs to dramatically reduce the number of policy parameters, therefore experience can be shared across other agents.
We evaluate MIXRTs on a range of challenging tasks in Spread and StarCraft II~\cite{samvelyan2019starcraft} environment.
Empirical results show that our learning architecture delivers simple explanations while enjoying competitive performance.
Specifically, MIXRTs find desirable optimal policies in easy scenarios compared to popular baselines like the widely investigated QMIX~\cite{rashid2018qmix} and QPLEX~\cite{wang2020qplex}.

The remaining paper is organized as follows.
In Section~\ref{Sec2}, we introduce basic concepts of MARL, SDTs, and related work.
In Section~\ref{Sec3} and \ref{Sec4}, we present the RTCs and the linear mixing architecture of individual RTCs, respectively.
In Section~\ref{Sec5}, we give experimental results of learning performance compared to the existent baselines.
The comprehensive interpretability of our model and results of the user study are given in Section~\ref{InterpretabilityInterpretability}.
Finally, we give concluding remarks in Section~\ref{Sec6}.

\section{Background and related works}\label{Sec2}

\subsection{Preliminaries}
\textbf{Dec-POMDP.}
The fully cooperative multi-agent task is generally modeled as a decentralized partially observable Markov decision process (Dec-POMDP)~\cite{oliehoek2016concise} that consists of a tuple $\left< \mathcal{S}, \mathcal{U}, \mathcal{P}, r, \mathcal{Z}, \mathcal{O}, n, \gamma \right>$, where $s\in \mathcal{S}$ describes the global state of the environment. 
At each time step, each agent $i \in \{1, ..., n\}$ only receives a partial observation $z \in \mathcal{Z}$ generated from an observation function $\mathcal{O}(s,i): S \times n \rightarrow \mathcal{Z}$, and chooses an action $u_i \in \mathcal{U}$  to formulate a joint action $\boldsymbol{u}:=\left[u_{i}\right]_{i=1}^{n} \in \mathcal{U}^{n}$.
This results in a transition to next state $s^{\prime} \sim P\left(s^{\prime} \mid s, \boldsymbol{u}\right)$. 
All agents share the same team reward signal $r(s,\boldsymbol{u}): \mathcal{S} \times \mathcal{U}^{n} \rightarrow \mathbb{R}$.
Furthermore, each agent learns its own policy $\pi_{i}\left(u_{i} \mid \tau_{i}\right): \mathcal{T} \times \mathcal{U} \rightarrow[0,1]$, which conditions on its action-observation history $\tau_{i} \in \mathcal{T}$,  and we define that $\boldsymbol{\tau}\in \boldsymbol{\mathcal{T}}$ is the joint action-observation history.
The goal is to find an optimal joint policy $\boldsymbol{\pi}=\left\langle\pi_{1}, \ldots, \pi_{n}\right\rangle$ to maximize the discounted
cumulative return $\sum_{t=0}^{\infty} \gamma^{t} r^{t}$, where $\gamma \in[0,1)$ is a discount factor.

\textbf{Multi-agent Q-learning in Dec-POMDP.} 
Q-learning~\cite{watkins1992q} is a classic tabular model-free algorithm to find the optimal joint action-value function $Q^*(s, \boldsymbol{u})=r(s, \boldsymbol{u})+\gamma \mathbb{E}_{s^{\prime}}[\max _{\boldsymbol{u}^{\prime}\in \mathcal{U}^n} $ $Q^*\left(s^{\prime}, \boldsymbol{u}^{\prime}\right)]$.
Multi-agent Q-learning approaches~\cite{sunehag2017value, son2019qtran, liu2023na2q} are generally based on the value decomposition extension of deep Q-learning~\cite{watkins1992q, mnih2015human}, where the agent system receives the joint action-observation history $\boldsymbol{\tau}$ and the joint action $\boldsymbol{u}$.
Given transition tuples $(\boldsymbol{\tau},\boldsymbol{u},r,\boldsymbol{\tau'})$ from the replay buffer $\mathcal{B}$, the network parameters $\mathit{\Theta}$ are learnt by minimizing the squared loss $\mathcal{L}( \mathit{\Theta})$ on the temporal-difference (TD) error $\sigma =y'- Q_{tot}(\boldsymbol{\tau}, \boldsymbol{u}; \mathit{\Theta})$, where $y'=r+\gamma \,\max _{\boldsymbol{u'}} Q_{tot}(\boldsymbol{\tau'}, \boldsymbol{u'}; \mathit{\Theta ' })$ is the target and $Q_{tot}(\boldsymbol{\tau}, \boldsymbol{u}; \mathit{\Theta})$ is used in place of $Q_{tot}(s, \boldsymbol{u}; \mathit{\Theta})$ due to partial observability. 
The parameters $\mathit{\Theta ' }$ from a target network are periodically copied from $\mathit{\Theta}$ and remain constant over multiple iterations.

\textbf{Centralized training and decentralized execution (CTDE).} 
In the CTDE fashion, during training the central controller can access the action-observation history of all agents and the global state, as well as the freedom to share all information between agents. 
While in the execution phase, each agent has its policy network to make decisions based on its individual action-observation history. 
Individual-global-max (IGM) is a common principle to realize effective CTDE, which asserts the consistency between joint and local greedy action selections by the joint value function $Q_{tot}$ and individual action-value function $Q_{i}(\tau_i, u_i)$:
\begin{gather}\label{Eq2}
\arg \underset{\boldsymbol{u}\in \mathcal{U}^n}{\max}\,Q_{tot}(\boldsymbol{\tau},\boldsymbol{u}) = \begin{pmatrix}
\arg \underset{u_1\in \mathcal{U}}{\max}\, Q_{1}(\tau _1,  u_1) \\
\vdots  \\
\arg\underset{u_n\in \mathcal{U}}{\max}\, Q_{n}(\tau_n, u_n)
\end{pmatrix},
\end{gather}
where $\bm{\tau}$ represents the joint action-observation history of all individual agents. 
Existing popular algorithms such as value decomposition networks (VDN)~\cite{sunehag2017value} and QMIX~\cite{rashid2018qmix} are based on CTDE, estimating the optimal $Q_{tot}$ via combining individual action-value function $Q_i$.
In VDN, the $Q_{tot}$ is calculated by summing the utilities of each agent as $Q_{tot}(\bm{\tau}, \boldsymbol{u}) = \sum_{i=1}^{n}Q_i(\tau_i, u_i)$. 
QMIX combines $Q_i$ through the state-dependent nonlinear monotonic function $f_s$: $Q_{tot}(\bm{\tau}, \boldsymbol{u}) = f_s(Q_1(\tau_1, u_1),... , Q_i(\tau_i, u_i), ..., Q_n(\tau_n, u_n))$, where $\frac{\partial f_s}{\partial Q_i}\geq 0, \forall i\in \{1, ..., n\}$.

\begin{figure*}[tb]
	\centering 
	\setlength{\abovecaptionskip}{4pt}
	\includegraphics[width=0.94\textwidth]{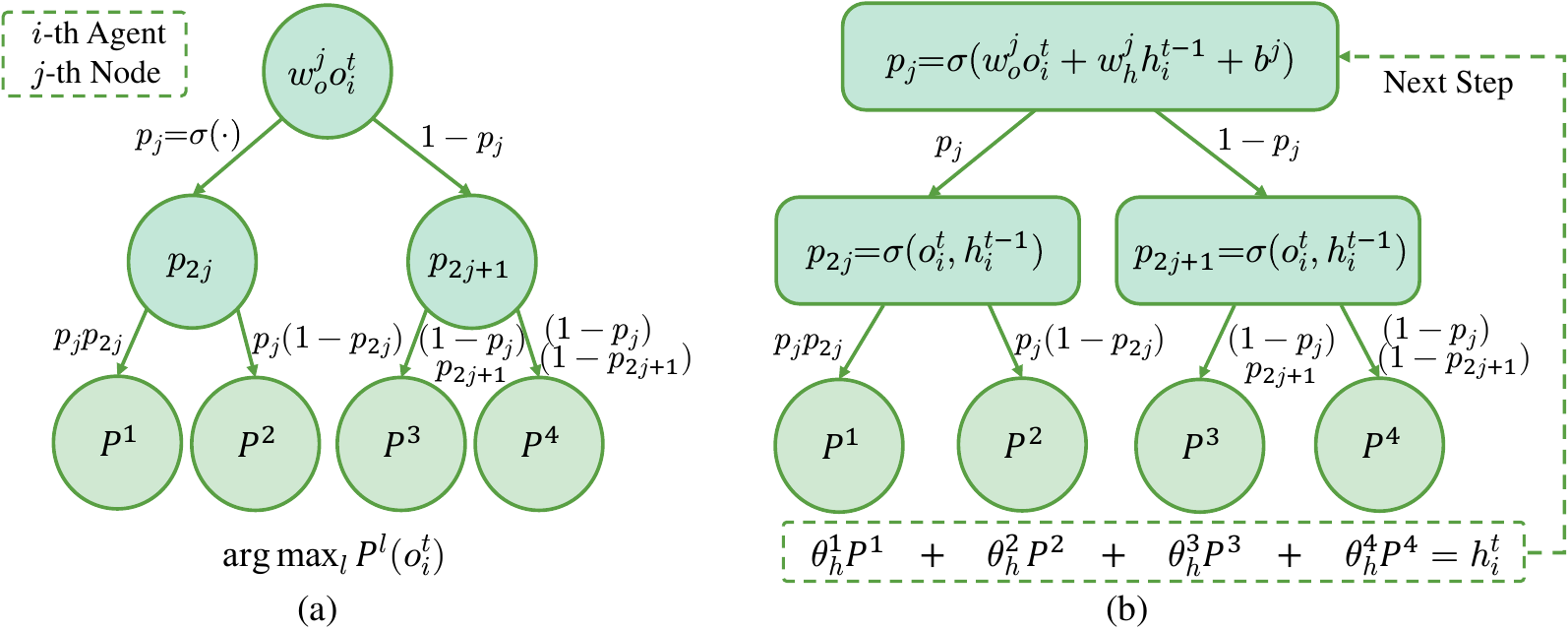}
	\caption{
		Examples of SDT and RTC.
		(a) A two-level SDT. 
		(b) Illustration of the process of a two-level RTC that receives the current individual observation $o^{t}_{i}$ and the previous hidden state $h^{t-1}_{i}$ as input at each timestamp.}
	\label{fig2}
\end{figure*}

\textbf{Soft decision trees (SDTs).} 
Differentiable decision trees~\cite{irsoy2012soft, frosst2017distilling} have been shown to own better expressivity compared to traditional hard decision trees. 
Especially, an SDT~\cite{frosst2017distilling} has favorable transparency and performance with a binary probabilistic decision boundary at each node in reinforcement learning tasks~\cite{coppens2019distilling, ding2020cdt}. 
As shown in Fig.~\ref{fig2}(a), it is different from the traditional growth of decision paths in that an SDT performs soft routing with different probabilities when given the fixed depth of the tree in advance.
Given an observation $o^t$, each inner node $j$ calculates the probability of traversing to its left child node by
\begin{gather}\label{Eq3}
p_j(o^t) = \sigma (w_o^j o^t+b^j),
\end{gather}
where $\sigma$ is a sigmoid function, and $w^j$ and $b^j$ are trainable parameters. 
The learned model consists of a hierarchy of decision filters that assign the input observation to a particular leaf node with a particular path probability $P^{l}(o^t)$, producing a probability distribution over the $\mathcal{U}$ action class.
Each leaf $l\in \mathit{Leaf Nodes}$ encodes relative action values $Q^{l} = \text{softmax}(\theta^{l})$, where $\theta^{l} \in \mathbb{R}^{\mathcal{U}\times 1}$ is a learnable parameter at the $l$-th leaf.
Following SDT, we select the action value $Q^{l_{\max}}$ with the largest probability leaf $l_{\max}$ at each timestamp $t$, where $l_{\max} = \arg\max_lP^{l}(o^t)$.

\subsection{Related works}
\textbf{Explainable reinforcement learning.}
In the reinforcement learning community, several prior works have investigated intrinsic methods aimed at deriving interpretable policies~\cite{silva2020optimization, irsoy2012soft,roth2019conservative}.
Intrinsic methods generally rely on using inherently interpretable models, such as classical decision trees and linear models, which directly represent the faithful decision-making process.
The conservative Q-improvement~\cite{roth2019conservative} introduces an interpretable decision-tree model that learns a policy as a hard decision tree for robot navigation tasks. 
It can balance the performance and simplicity of the policy model only when adding new splits in a task.
Differentiable decision trees methods~\cite{silva2020optimization,pace2021poetree} allow for a gradient update over Q-learning and policy gradient algorithms by replacing the Boolean decision with a sigmoid activation function at the decision node, which improves the performance while affording interpretable tree-based policies.
These methods fix their structure before the learning process rather than incrementally adding nodes, where the leaf nodes represent single features using discretization techniques to enhance interpretability.
Instead of constructing a univariate decision tree, the cascading decision tree (CDT)~\cite{ding2020cdt}, an extension based on an SDT with multivariate decision nodes, applies representation learning to decision paths to achieve richer expressive capabilities.
The prior works closest to ours are SDT~\cite{frosst2017distilling} and CDT~\cite{ding2020cdt}, whose policies are generated by the routing probabilities at each leaf node to make decisions.
Our method differs in that 1) we introduce the recurrency into SDTs for capturing the long-term condition in partially observable tasks, 2) we explicitly present the graphical decision process with a smaller depth by considering all variate at each node instead of univariate features, and 3) we interpret the role each agent plays in a team by visualizing the linear credit assignment.

\textbf{Cooperative multi-agent reinforcement learning.}
Dating back to the early stage of MARL, independent Q-learning~\cite{tan1993multi} is a common method, in which each agent optimizes its policy that is independent of others.
In this approach, agents observe only local information, execute their actions, and receive rewards individually, effectively treating other agents as part of the environment.
However, this method lacks the theoretical convergence guarantees of standard Q-Learning due to the non-stationarity introduced by changes in other agents' policies.
By contrast, CTDE~\cite{oliehoek2008optimal, kraemer2016multi} is an advanced paradigm for cooperative MARL tasks, which allows each agent to learn a joint action-value function via credit assignment mechanisms.
VDN~\cite{sunehag2017value} factorizes the joint action-value function into a linear summation over individual agents.
To address the VDN's limitation of ignoring additional global state information, QMIX~\cite{rashid2018qmix} utilizes a non-linear mixing network augmented with a state-dependent bias to estimate the joint action-value function more effectively.
QTRAN~\cite{son2019qtran} further relaxes the constraints on the greedy action selections between the joint and individual functions using two soft regularizations.
Different from QTRAN which loses the exact IGM consistency, QPLEX~\cite{wang2020qplex} uses a dueling network architecture to represent both joint and individual action-value functions to guarantee the IGM property.

Our method distinguishes itself from these representative value decomposition methods in several key aspects: 1) Our method builds upon soft decision trees, which not only provide intrinsic interpretability for visualization but also preserve historical information through a recurrent structure.
2) Our mixing trees module linearly factorizes the joint action value into individual action values, achieving high scalability by fully implementing the IGM principle.
This interpretable model enables us to explain not only the explicit behavior of each agent but also the behaviors of different roles within the team. 
3) Our proposed method requires only lightweight linear reasoning while achieving competitive performance across a series of collaborative tasks.

\section{Recurrent Tree Cells}\label{Sec3}
In this section, we propose RTCs,  which introduce recurrency into SDTs to encode historical information and enhance the fidelity of the Q-value through the ensemble technique. 
First, we propose an RTC that receives the current individual observation and relies on historical information to capture long-term dependencies in partially observable tasks.  
Then, we utilize a linear combination of multiple RTCs via an ensemble framework to yield high performance and reduce the variance of the model, all while retaining simplicity and interpretability.

\subsection{Recurrent Tree Cell} 
The complexity of neural networks brings barriers to understanding, verifying, and trusting the behaviors of agents, as it has complex transformations and non-linear operators.
To relieve this dilemma, the SDT offers an effective way to interpret the decision-making pathways by visualizing decision nodes and associated probabilities.
As depicted in Fig.~\ref{fig2}(a), an SDT shows an univariate differentiable tree with a probabilistic decision boundary at each filter. 
However, the SDT lacks explicit mechanisms for deciphering the underlying state of POMDP in sequential decision-making tasks over longer timescales.
This limitation can degrade performance by estimating the Q-value from incomplete observations rather than the global state.
Indeed, by leveraging recurrency to a deep Q-network~\cite{hausknecht2015deep}, a recurrent neural network~(RNN) can effectively capture long-term dependencies conditioning on their entire action-observation history.  
Inspired by advanced recurrent architectures, we first introduce recurrence into an SDT and propose a recurrent tree cell (RTC) to capture long-term dependencies.
As shown in Fig.~\ref{fig2}(b), an RTC receives the current individual observation $o^{t}_{i}$ and the previous embedding $h^{t-1}_{i}$ as input at each time step, extracting hidden state information for each agent $i$.

For an RTC, each non-leaf node learns a linear filter to traverse to its left child node with the probability as
\begin{gather}\label{Eq4}
p_j(o_{i}^t, h_i^{t-1}) = \sigma (w^{j}_o o_{i}^t + w^j_h h_i^{t-1} + b^j),
\end{gather}
where $w^j_o$ and $w^j_h$ are learnable parameters for the observation $o^{t}_{i}$ and the previous hidden state $h^{t-1}_{i}$, respectively, and $b^j$ is a learnable bias.
Similarly, the probability of traversing to the right child is $1 - p_j(o_{i}^t, h_i^{t-1})$.

We now discuss how to model the probability that an observation reaches a selected leaf $l$.
Let $\left [ l\swarrow j \right ]$ denotes a boolean event $\in \left \{ 0, 1 \right \}$, indicating whether $j$ passes the left-subtree to reach the leaf node $l$. 
The probability $P^l$ that the observation $o_{i}^t$ reaches $l$ for each agent $i$ is given by
\begin{equation}\label{Eq5}
\begin{aligned}
P^l(o_{i}^t, h_i^{t-1}) = \prod_{j\in route(l)} \left\{ p_j(o_{i}^t, h_i^{t-1})^{\left [ l\swarrow j \right ]}\,   \right.\quad\\
\left. (1-p_j(o_{i}^t, h_i^{t-1}))^{1-\left [ l\swarrow j \right ]}   \right\},
\end{aligned}
\end{equation}
where $p_j$ is the probability of going from a current node $j$ to its left child node $2j$, as illustrated in Fig.~\ref{fig2}.

With the target distribution of the tree, we measure the current hidden state $h_i^t$ of the leaf by combining the probability values of each leaf with a scalar weight $\theta^l_h$, and provide a vector $w_q$ that serves this tree to capture the action-observation value as
\begin{equation}\label{Eq6}
\begin{aligned}
h_i^t = & \sum_{l\in \mathit{Leaf Nodes}}\theta^l_hP^l(o_{i}^t, h_i^{t-1}), \\
Q_i(\tau_i, \cdot)  = &~~~~~~w_q^\top h_i^t,
\end{aligned}
\end{equation}
where $P^l$ is the overall path probability along the root to leaf $l$.
Different from the leaf nodes of the SDT, each learnable parameter $\theta^l_h\in \mathbb{R}$  calculates the hidden state $h_i^t$ by weighting with $P^l$, and $w_q\in \mathbb{R}^{1\times \mathcal{U}}$ is a training parameter to transform the hidden state $h_i^t$ into the action distribution in an RTC.  
Finally, during decentralized execution, each agent $i$ selects an action $u_i^{t}$ using $\varepsilon$-greedy strategy with respect to its estimated $Q_i(\tau_i, u_i)$.
This decision process retains the simplicity and interpretability of the model, as it reveals which dimensions of the observation influence the action distribution during the inference.

\subsection{Ensemble of Recurrent Tree Cells} 
SDT-based methods exhibit several appealing properties in multivariate tree structures, such as ease of tuning, robustness to outliers, and good interpretability~\cite{ding2020cdt}.
Since all input features are used for each node in a multivariate setting, a single SDT may lack expressiveness and output predictions with high variance.
It is well known that an ensemble of models can reduce the variance component of the estimation error~\cite{zhang2012ensemble,hazimeh2020tree}.
To relieve the above tension while maintaining the interpretability, we linearly combine multiple RTCs with the variance-optimized bagging (Vogging) approach based on the bagging technique~\cite{derbeko2002variance}.
Based on the Vogging ensemble mechanism, the individual value function for the agent $i$ can be represented as
\begin{gather}\label{eq6}
Q_i(\tau_i, \cdot)  = w_{q,(1)}^\top\,h_{i,(1)}^{t}  +  w_{q,(2)}^\top\,h_{i,(2)}^{t} +...+ w_{q,(H)}^\top\,h_{i,(H)}^{t},
\end{gather}
where $H$ is the size of ensemble RTCs, $w_{q,(1)},  w_{q,(2)}, ...,  w_{q,(H)}$ are learnable parameters used to optimize the linear combination of the $H$ trees for improving the expressiveness and reducing the prediction variance of RTCs, and the hidden state $h^t_i \in \mathbb{R}^{H\times 1}$ is rewritten as the vector $[h_{i,(1)}^{t}, h_{i,(2)}^{t}, ..., h_{i,(H)}^{t}]^\top$ to represent the history record, i.e., $h_i$ is a representation of the history $\tau_i$. 
Finally, RTCs produce a local policy distribution as a Q-function for each agent and execute an action by sampling this distribution.
Here, the loss function of RTCs working on a single agent is based on Q-learning~\cite{watkins1992q,mnih2015human}, updating $Q_i$ as
\begin{gather}\label{xxxxsadasd}
\mathcal{L} = \mathbb{E}\left [\left( r_i + \gamma\max_{u'} Q_i(\tau_i',u') -Q_i(\tau_i, u) \right)^2\right ],
\end{gather}
where $r_i$ is the reward received by agent $i$. 
However, in multi-agent systems, all agents only share a team reward $r$, and therefore we explore the value decomposition in Section~\ref{Sec4}.

\begin{remark}
The functions used for all sub-modules in the RTCs are linear, which both preserves the interpretability of the model and simplifies its structure.
It is important to note that as the model becomes more complex, interpretability diminishes due to the loss of simplicity.
Since shallow trees can be easily described and even be implemented manually, it is acceptable that linearly combining trees of depth $2$ or $3$ achieves acceptable performance and stability while only sacrificing a minimal degree of interpretability.
In practice, a moderate number of ensemble trees $H$ (e.g., ranging from $16 \sim 64$) is sufficient to obtain efficient performance, and additional ablation results are provided in Section~\ref{ablation}.
\end{remark}

\section{Mixing Recurrent SDTs Architecture}\label{Sec4}
\begin{figure*}[tb]
	\centering 
	\subfigbottomskip=0pt 
	\setlength{\abovecaptionskip}{0pt}
	\subfigcapskip=-0pt 
	\includegraphics[width=0.99\textwidth]{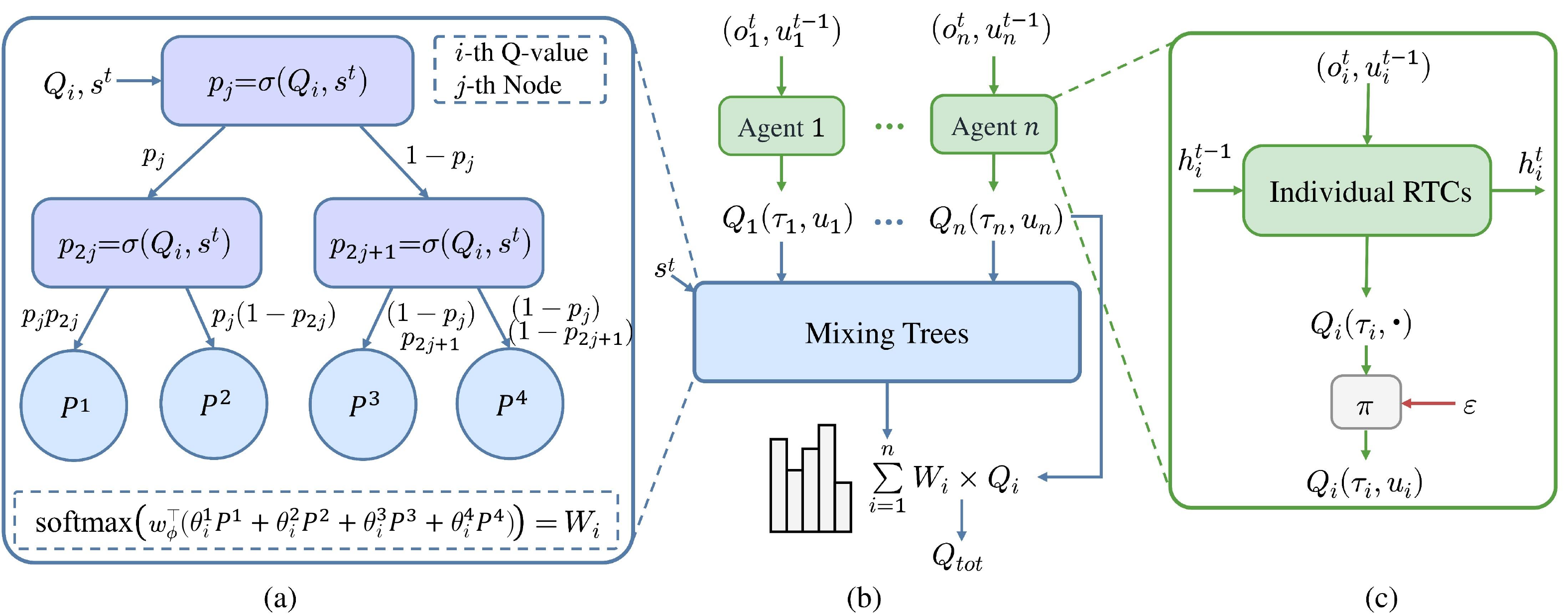}
	\caption{MIXRTs architecture. (a) Diagram of the structure of the mixing tree with depth $2$. 
		(b) In the overall MIXRTs architecture, we finally obtain the joint $Q_{tot}$ value via a linear combination of the individual action-value functions. 
		(c) Individual RTCs for each agent.
	}
	\vskip -0.1in
	\label{fig3}
\end{figure*}
In this section, we propose a novel method called \textit{MIXing Recurrent soft decision Trees} (MIXRTs), which can represent a much richer class of action-value functions analogous to advanced CTDE algorithms.
The overall architecture of MIXRTs is illustrated in Fig.~\ref{fig3}, including two main components as follows: (i) an ensemble of RTCs as an individual action-value function for each agent, and (ii) a mixing component similar to the ensemble of RTCs in which a joint action-value function $Q_{tot}$ is factorized into the individual action-value function $Q_{i}$ of each agent $i$ under the IGM constraint.
The two components are trained in a centralized manner and each agent uses its own factorized individual action-value function $Q_i$ to take actions during decentralized execution.
Each component is elaborated on next.

\subsection{Individual RTCs for the Action Value}
For each agent, the individual action-value function can generally be represented by an ensemble of RTCs, where different agents are mutually independent RTCs.
However, as the number of agents increases, the joint action-observation space grows exponentially, leading to a large number of learnable parameters that may confuse understanding the decision process of the model.
Inspired by parameters sharing~\cite{sunehag2017value}, we utilize the ensemble of RTCs with shared parameters among agents to improve the learning efficiency.
This operation can also give rise to the concept of agent invariance and help avoid the lazy agent problem.
To keep agent invariance, we incorporate role information into each agent via a one-hot encoding of its identity, which is concatenated with the corresponding observation at the root layer.
The architecture, with shared weights and information channels across agent networks, satisfies agent invariance with the identical policy, thereby enhancing simplicity and interpretability with fewer parameters.
For each agent $i$, the individual value is represented by an ensemble of RTCs that takes the current local observation $o^{t}_{i}$, the previous hidden state $h^{t-1}_i$ and the previous action $u^{t-1}_i$ as inputs, and then outputs the individual action value $Q_i(\tau_i, u_i)$.

\subsection{Mixing Trees Architecture}
The mixing trees architecture follows an interpretable Q-value decomposition that connects the local and joint action-value functions via feed-forward ensemble trees, which is similar to the ensemble of RTCs but without embedding the history information.
During the centralized training process, it incorporates the information of global state $s_t$ into individual action-value functions and produces the weights of the local action-value in the joint action-value $Q_{tot}$.
The main assumption is that the joint action-value function can be additively decomposed into value functions across agents.
However, unlike VDN or QMIX, we assume that the joint action-value function can be approximated as a linear weighting of the individual Q-values
\begin{gather}\label{Eq9}
Q_{tot}(\bm{\tau},\boldsymbol{u}) \approx \sum_{i=1}^{n}W_i Q_i(\tau_i, u_i),
\end{gather}
where $\mathbf{\bm{\tau}}$ and $\boldsymbol{u}$ are joint action-observation and joint action, respectively, and we force the assignment of positive credit weights $W_i > 0$. 
The linear weighting is chosen to provide a reliable interpretation of the relationships among agents.
In general, the agent assigned a higher credit assignment with a larger weight $W_i$ makes a greater contribution to the team.

In detail, we employ the interpretable RTCs to obtain weights $W_i$ for each agent $i$, which establish relations from the individuals to the global.
Each filter of RTCs takes the individual action value $Q_i$ and the global state $s^t$ as inputs, and calculates the probability of transitioning to the left child node as
\begin{gather}\label{Eq10}
p_j(Q_i, s^t) = \sigma (w^j_q Q_i+w^j_s s^t+b^j),
\end{gather}
where $w^j_q$, $w^j_s$, and $b^j$ are learnable parameters for each node of RTCs.
After the inference process, we obtain the joint probability distributions $P^{l}$ at the leaf layer. 
To achieve stable weights, we employ an ensemble mechanism with $H$ trees, resulting in
\begin{equation}\label{Eq11}
\begin{aligned}
\phi _i = & \sum_{l\in \mathit{Leaf Nodes}}^{}\theta^l_iP^l(Q_i, s^t), \\
W_i = & \frac{\exp( \sum_{k=1}^{H} w _{\phi,(k)}^\top\,\phi _{i,(k)}  )}{\sum_{i=1}^{n}\exp( \sum_{k=1}^{H} w _{\phi,(k)}^\top\,\phi _{i,(k)} )},
\end{aligned}
\end{equation}
where $\theta^l_i\in \mathbb{R}$ and $w_\phi\in \mathbb{R}^{1\times n}$ are learnable parameters in one tree.
The mixing weight $W_i$ is ensured to be positive through the use of softmax operation, which enforces monotonicity between $Q_{tot}$ and each $Q_i$. 
Here, the goal of the mixing architecture is not only to achieve efficient value decomposition but also to provide an intuitive interpretation that highlights the importance of observation attributes and the weights of each agent at different time steps within an episode.
Further, Theorem~\ref{theorem1} provides a theoretical analysis showing that MIXRTs guarantee monotonicity by imposing a simple constraint on the relationship between  $Q_{tot}$ and each $Q_i$, which is sufficient to satisfy Eq.~\ref{Eq2}.
This allows each agent $i$ to participate in a decentralized execution solely by choosing greedy actions.

\begin{theorem}\label{theorem1}
    Let $Q_{tot}(\bm{\tau}, \boldsymbol{u})$ be a joint action-value function that is factorized by a set of individual action-value functions $Q_{i}(\tau_i, u_i)$ as $Q_{tot}(\bm{\tau},\boldsymbol{u}) \approx \sum_{i=1}^{n}W_i Q_i(\tau_i, u_i)$, where $W_i>0$ and $n$ is the number of agents.
    Then, we have $\boldsymbol{u^\ast } = \arg \max_{\boldsymbol{u}\in \mathcal{U}^n} Q_{tot}(\bm{\tau}, \boldsymbol{u})$ that satisfies the IGM principle. 
\end{theorem}

\begin{proof}
	For an arbitrary factorizable $Q_{tot}(\bm{\tau}, \boldsymbol{u})$, we take $\boldsymbol{u^\ast } = \arg \max_{\boldsymbol{u}\in \mathcal{U}^n} Q_{tot}(\bm{\tau}, \boldsymbol{u})$. 
	Recall that $\boldsymbol{u^\ast }:=\left[u^\ast_{i}\right]_{i=1}^{n} \in \mathcal{U}^{n}$ and $u^\ast _i = \arg \max_{u_i\in \mathcal{U}} Q_{i}(\tau_i, u_i)$. 
	After the softmax operation, the weight of each individual action-value function satisfies $W_i> 0$. For $\forall i \in \{1, 2, ..., n\}$,  we have monotonicity, i.e.,
	\begin{equation}\label{eq11dasda1}
	\frac{\partial Q_{tot} }{\partial Q_i} > 0.
	\end{equation}
	Thus, for any $\left[u_{i}\right]_{i=1}^{n}$ and the mixing trees function $Q_{tot}(\cdot)$ with $n$ weights, the following holds 
	\begin{equation*}
	\begin{aligned}
	& \ \ \ \ \ \ Q_{tot}(Q_1(\tau_1, u_1),Q_2(\tau_2, u_2), \cdots ,Q_n(\tau_n, u_n))\\
	&:= W_1 Q_{1}(\tau _1,  u_1)+ W_2 Q_2(\tau_2, u_2)+ \cdots + W_n Q_n(\tau_n, u_n)\\
	& \leq W_1 Q_{1}(\tau _1,  u^\ast_1)+ W_2 Q_2(\tau_2, u^\ast_2)+ \cdots + W_n Q_n(\tau_n, u^\ast_n).\\
	\end{aligned}
	\end{equation*}
	Thus, according to the $\arg \max$ operator, we have the maximum of the joint action value
	\begin{equation*}
	\begin{aligned}
	& \, \, \, \, \ \ \, \ \underset{\boldsymbol{u}\in \mathcal{U}^n}{\max}\, Q_{tot}(\bm{\tau}, \boldsymbol{u})\\
	& :=\underset{\boldsymbol{u}\in \mathcal{U}^n}{\max}\, Q_{tot}(Q_1(\tau_1, u_1),Q_2(\tau_2, u_2), \cdots ,Q_n(\tau_n, u_n))\\
	&\ 	=Q_{tot}(Q_{1}(\tau _1,  u^\ast_1),  Q_{2}(\tau _2,  u^\ast_2), \cdots ,  Q_n(\tau_n, u^\ast_n)).
	\end{aligned}
	\end{equation*}
	Hence, we have $\boldsymbol{u^\ast } = \arg \max_{\boldsymbol{u}\in \mathcal{U}^n} Q_{tot}(\bm{\tau}, \boldsymbol{u})$ that satisfies the IGM principle, and the assumed mixing trees provide universal function approximation weights by Eq.~(\ref{Eq9}).
\end{proof}

All parameters $\mathit{\Theta}$ in MIXRTs are learned by sampling a batch of transitions $b$ from the buffer $\mathcal{B}$ and minimizing the following expected squared TD error loss as
\begin{gather}\label{Eq13}
\mathcal{L}(\mathit{\Theta }) = \mathbb{E}_{(\mathbf{\bm{\tau}},\boldsymbol{u},r,\mathbf{\bm{\tau}}')\in b}\left [( y' -Q_{tot}(\mathbf{\bm{\tau}}, \boldsymbol{u}; \mathit{\Theta }) )^2\right ],
\end{gather}
where the target value $y'$ is obtained using Double DQN~\cite{hasselt2010double,van2016deep} and is estimated as $y'=r+\gamma \overline{Q}_{tot}(\bm{\tau'}, \arg \max_{\bm{u'}\in \mathcal{U}^n}{Q}_{tot}(\mathbf{\bm{\tau'}}, \bm{u'}; \mathit{\Theta'}))$.
Here, $\mathit{\Theta '}$ denotes all parameters of the target network, which are periodically copied from $\mathit{\Theta}$.
Since Eq.~(\ref{Eq2}) holds, we maximize $Q_{tot}$ in a linear fashion to achieve competitive performance without resorting to more complex non-linear methods.

\begin{table*}[tb]
	\caption{The StarCraft multi-agent challenge benchmark.}\label{table3}
	\centering
	\fontsize{9}{11}\selectfont 
	\begin{tabular}{l|cccc}
		\hline
		\hline
		Scenarios Type&Map &Ally Units & Enemy Units & Total  Steps \\
		\hline
		\multirow{3}{*}{Easy} &3m& 3 Marines& 3 Marines&$1050K$\\
		&8m & 8 Marines& 8 Marines&$1500K$\\
		&2s3z & 2 Stalkers, 3 Zealots& 2 Stalkers, 3 Zealots&$1500K$\\
		&2s\_vs\_1sc&  2 Stalkers&1 Spine Crawler&$2M$\\
		\hline
		\multirow{3}{*}{Hard} &5m\_vs\_6m& 5 Marines& 6 Marines&$2M$\\
		&3s5z &3 Stalkers, 5 Zealots&3 Stalkers, 5 Zealots&$2M$\\
		&8m\_vs\_9m& 8 Marines& 9 Marines&$2M$\\
		\hline
		\multirow{3}{*}{Super Hard} &6h\_vs\_8z & 6 Hydralisks& 8 Zealots&$5M$\\		&\multirow{2}{*}{MMM2}& 1 Medivac, 2 Marauders,&1 Medivac, 	3 Marauders,&\multirow{2}{*}{$2M$ }\\
		& &and  7 Marines&and 8 Marines&\\
		\hline
		\hline
	\end{tabular}
\end{table*}

\section{Experiments}\label{Sec5}
In this section, we evaluate MIXRTs on two representative benchmarks as our testbed: the Multi-agent Particle Environment (MPE)~\cite{lowe2017multi} and StarCraft Multi-Agent Challenge (SMAC)~\cite{samvelyan2019starcraft}.
The goal of our experiments is to evaluate the performance and demonstrate the interpretability of MIXRTs.
We compare our method with widely investigated algorithms, including VDN~\cite{sunehag2017value}, QMIX~\cite{rashid2018qmix}, QTRAN~\cite{son2019qtran} and QPLEX~\cite{wang2020qplex}, since these are advanced value-based methods that train decentralized policies in a centralized fashion.
It is important to note that our focus is on balancing model interpretability and learning performance, rather than blindly beating state-of-the-art baselines.
Additionally, we compare MIXRTs with existing interpretable models SDTs~\cite{frosst2017distilling} and CDTs~\cite{ding2020cdt}, which decompose a multi-agent task into a set of simultaneous single-agent tasks~\cite{tan1993multi, tampuu2017multiagent}.
Further, we perform the sensitivity analysis on two key hyperparameters: the tree depth and the number of trees.
Finally, we concretely present the interpretability of MIXRTs regarding the learned tree model at SMAC, aiming to present an easy-to-understand decision-making process from the captured knowledge of tasks.

\subsection{Environmental description}\label{Sec5.13}
\textbf{Spread environment in MPE.} 
We select the commonly used Spread environment in the MPE, where the objective for a set of $n$ agents is to navigate toward randomly allocated positions marked by $n$ landmarks while evading any inter-agent collisions.
The ideal strategy involves each agent exclusively claiming a single landmark, achieving an optimal spatial distribution.  
This necessitates advanced coordination among agents, as they must predict the targets that their counterparts will likely claim and adjust their actions accordingly to occupy the remaining landmarks.
The observation space for a given agent $i$ is represented by a vector that includes the agent's velocity and absolute positional coordinates, as well as the relative positions of all other agents and landmarks. 
Each agent can move in one of four cardinal directions or remain stationary. 
The collective reward for the agents is quantified as the negative sum of the minimal distances between each landmark and the nearest agent.
To discourage inter-agent collisions, an additional penalizing factor is incorporated into the reward computation, which is defined following the MPE~\cite{lowe2017multi}.
We conduct experiments with $3$, $4$, and $5$ agents and set the number of training steps to $4M$ to ensure convergence.

\textbf{StarCraft II.}
This environment is based on StarCraft II unit micro-management tasks (SC2.4.10 version). 
We consider combat scenarios where the enemy units are controlled by a built-in AI with the \textit{difficulty=7} setting, and each allied unit is controlled by the decentralized agents using RL.
During battles, the agents seek to maximize the damage dealt to enemy units while minimizing damage received, which requires a range of skills.
We evaluate our method on a diversity of challenging combat scenarios, and Table~\ref{table3} presents a brief introduction to these scenarios with symmetric/asymmetric agent types and varying agent numbers.
To better interpret MIXRTs in such complex tasks, we provide a detailed description of SMAC~\cite{samvelyan2019starcraft} in detail, including the observations, states, actions, and rewards settings.

\textit{Observations and states.} At each time step, each agent receives local observations within its field of view, including the following attributes for both allied and enemy units: distance, relative coordination in the horizontal axis (relative X), relative coordination in the vertical axis (relative Y), health, shield, and unit type.
All Proto's units possess shields that serve as a defense mechanism against attacks and can regenerate if no further damage is incurred.
Notably, Medivacs, being healer units, play a crucial role in maintaining the health of agents during battles.
Unit types are used to distinguish different kinds of units on heterogeneous scenarios (e.g., 2s3z and MMM2 in the experiment).
Agents can only observe other units if they are alive and within their line of sight range, which is set to $9$.
If a unit (for both allies and enemies) feature vector is reset to zeros, it indicates either the unit’s death or its invisibility due to being outside another agent’s sight range.
The global state, containing information about all units on the map, is available only to agents during centralized training.
Finally, all features, including observations and the global state, are normalized by their maximum values.

\textit{Action space.} Each unit takes an action from a discrete action set: no operation (no-op), stop, move~[direction], and attack~[enemy id].
Agents can move with a fixed step size in four directions: north, south, east, and west, and are permitted to execute the attack~[enemy id] action only when the enemy is within its shooting range.
Note that a dead unit can only take the no-op action, while a living unit cannot select no-op.
Finally, depending on different scenarios, the maximum number of operations that a living agent can perform typically ranges from $7$ to $70$.

\textit{Rewards.}
The objective is to maximize the win rate for each battle scenario.
At each time step, agents receive a shaped reward based on the hit-point damage dealt and enemy units killed.
Additionally, agents receive a bonus of $10$ after killing each enemy, and a $200$ bonus when killing all enemies, which is consistent with the default reward function of the SMAC.
To ensure consistency across different scenarios, reward values are scaled so that the maximum cumulative reward is around $20$.

\begin{table}[tb]
	\caption{Hyperparameters settings of value-based algorithms.}\label{table4}
	\centering
	\fontsize{9}{10}\selectfont 
	\begin{tabular}{l|cc}
		\hline
		\hline
		Method & Hyperparameter/Description & Value \\
		\hline
		\multirow{10}{*}{Common} &Difficulty of the game & $7$ \\
		&Evaluate Cycle &$5000$\\
		&Target Update Cycle&$200$\\
		&Number of the epoch to evaluate agents&$32$ \\
		&Optimizer&RMSprop\\
		&Discount Factor $\gamma $ &$0.99$\\
		&Batch Size& $32$\\
		&Buffer Size& $5000$\\
		&Anneal Steps for $\varepsilon $& $50000$\\
		&	Learning Rates& $0.0005$\\
		\hline
		VDN &Agent RNN Dimension& $64$\\
		\hline
		\multirow{2}{*}{QMIX} &Agent RNN Dimension& $64$\\
		&Mixing Network Dimension& $64, 32$\\
		\hline
		\multirow{4}{*}{QTRAN} &Agent RNN Dimension&$ 64$\\
		&Mixing Network Dimension&$ 64$\\
		&Lambda-opt& $1.0$\\
		&Lambda-nopt& $0.1$\\
		\hline
		\multirow{3}{*}{QPLEX} &Agent RNN Dimension& $64$\\
		&Mixing Network Dimension& $64, 32$\\
		&Attention Embedded Dimension& $64, 64, 64$\\
		\hline
		\multirow{1}{*}{I-SDTs}&Agent Trees Depth& $3$\\
		\hline
		\multirow{3}{*}{I-CDTs} &Intermediate Variables Dimension& $32$\\
		&Intermediate Feature Depth& $3$\\
		&Decision Depth& $3$\\
		\hline
		\multirow{2}{*}{I-RTCs} &Agent Trees Ensemble Dimension& $32$\\
		&Agent Trees Depth& $3$\\
		\hline
		\multirow{4}{*}{MIXRTs} &Agent Trees Ensemble Dimension& $32$\\
		&Agent Trees Depth& $3$\\
		&Mixing Trees Ensemble Dimension& $16$\\
		&Mixing Trees Depth& $3$\\
		\hline
		\hline
	\end{tabular}
\end{table}

\subsection{Experimental Setup}\label{Sec5.1}

We compare our method with widely investigated value decomposition baselines, including VDN~\cite{sunehag2017value}, QMIX~\cite{rashid2018qmix}, QTRAN~\cite{son2019qtran}, and QPLEX~\cite{wang2020qplex}, based on an open-source implementation of these algorithms~\footnote{\url{https://github.com/oxwhirl/pymarl}}.
Besides, we compare with existing interpretable models of SDTs~\cite{frosst2017distilling} and CDTs~\cite{ding2020cdt}, which decompose the problem into a set of simultaneous single-agent problems via the independent Q-learning~\cite{tan1993multi, tampuu2017multiagent} structure, referred to as I-SDTs and I-CDTs, respectively.
Similarly, to ensure fair comparisons, we also implement RTCs with independent Q-learning to verify the reliability of the module, referred to as I-RTCs.
The hyperparameters and environmental settings of these algorithms are consistent with their source codes, adhering to the SMAC configuration~\cite{samvelyan2019starcraft}.
More details can be found in Table~\ref{table4}.

\begin{figure*}[tb]
	\centering 
	\subfigbottomskip=0pt 
	\subfigcapskip=-5pt 
	\setlength{\abovecaptionskip}{5pt}
	\includegraphics[width=.99\textwidth]{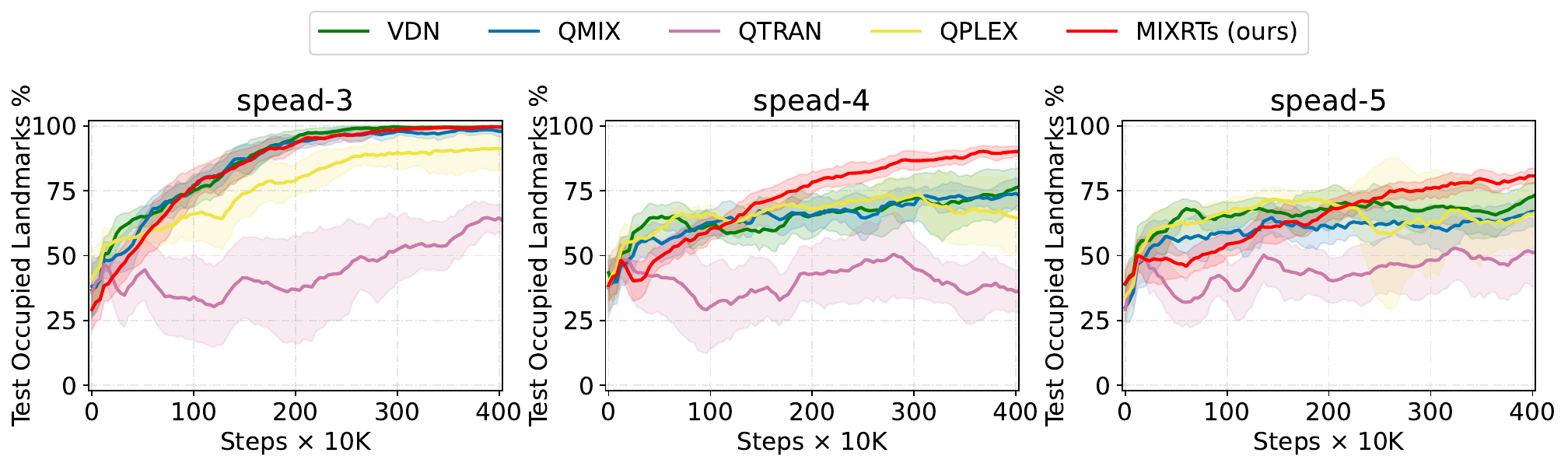}
	\caption{Median test occupied landmarks \% across different numbers of agents in the Spread environment.}
	\label{figfafasf15}
\end{figure*}

\begin{table*}[tb]
	\caption{The number of parameters needs to be learned for different algorithms on different scenarios.}\label{table1}
	\centering
	\fontsize{9}{12}\selectfont 
	\begin{tabular}{l|ccc|ccc|cc}
		\hline
		\hline
		Method & 3m& 8m  & 2s3z & 5m\_vs\_6m & 3s5z &8m\_vs\_9m& MMM2& 6h\_vs\_8z\\
		\hline
		VDN & $28,297$ & $32,462 $&$31,883$ &$30,412$ &$35,534$  &$32,911$ &$39,250$ &$ 32,206$ \\
		QMIX & $46,058$ & $83,663 $& $ 67,628$ &$61,933$& $ 95,951  $ &$86,224$ &$124,179$  & $73,871$\\
		QTRAN & $58,495 $& $79,860 $& $ 72,021 $&$68,102$ & $89,076$ & $82,097$ &  $107,904$& $76,020$\\
		QPLEX & $275,669 $& $587,144 $& $ 447,491 $&$404,164$ & $688,520$ & $615,241$ &  $955,704$& $508, 764$\\
		\hline
		MIXRTs (ours)& $20,880$& $37,040$&$34,448 $  & $28,752$ &$48,560$ & $38,592$ &$62,736 $& $35,440 $\\
		\hline
		\hline
	\end{tabular}
\end{table*}

\begin{figure*}[tb]
	\centering 
		\subfigbottomskip=0pt 
	\subfigcapskip=-0pt 
	\includegraphics[width=0.99\textwidth]{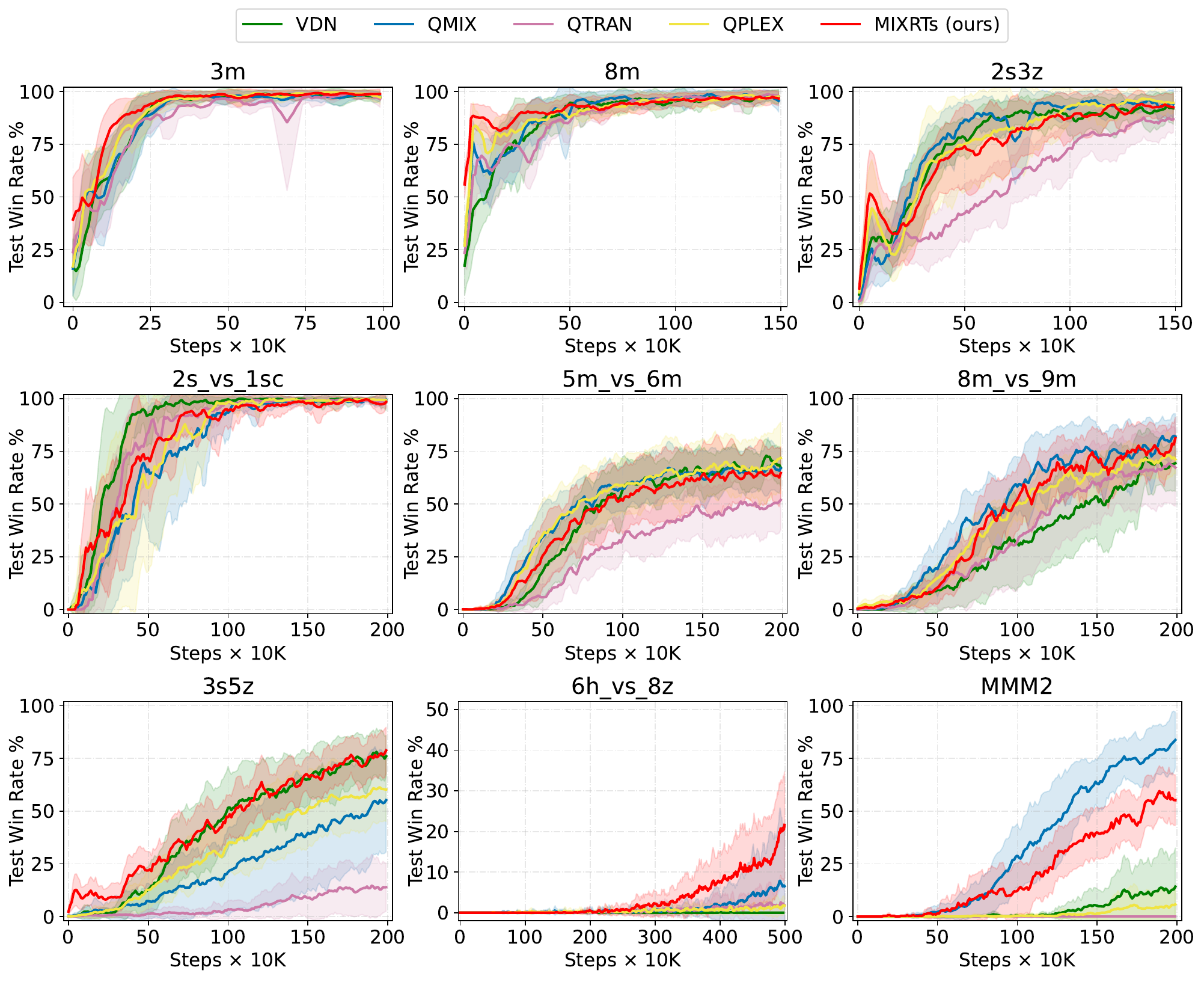}
	\caption{Median test win rates \% for easy, hard and super-hard scenarios of SMAC.}
	\label{fig4}
\end{figure*}

\begin{figure*}[tb]
	\centering 
	\subfigbottomskip=0pt 
	\subfigcapskip=-0pt 
	\setlength{\abovecaptionskip}{5pt}
	\includegraphics[width=.99\textwidth]{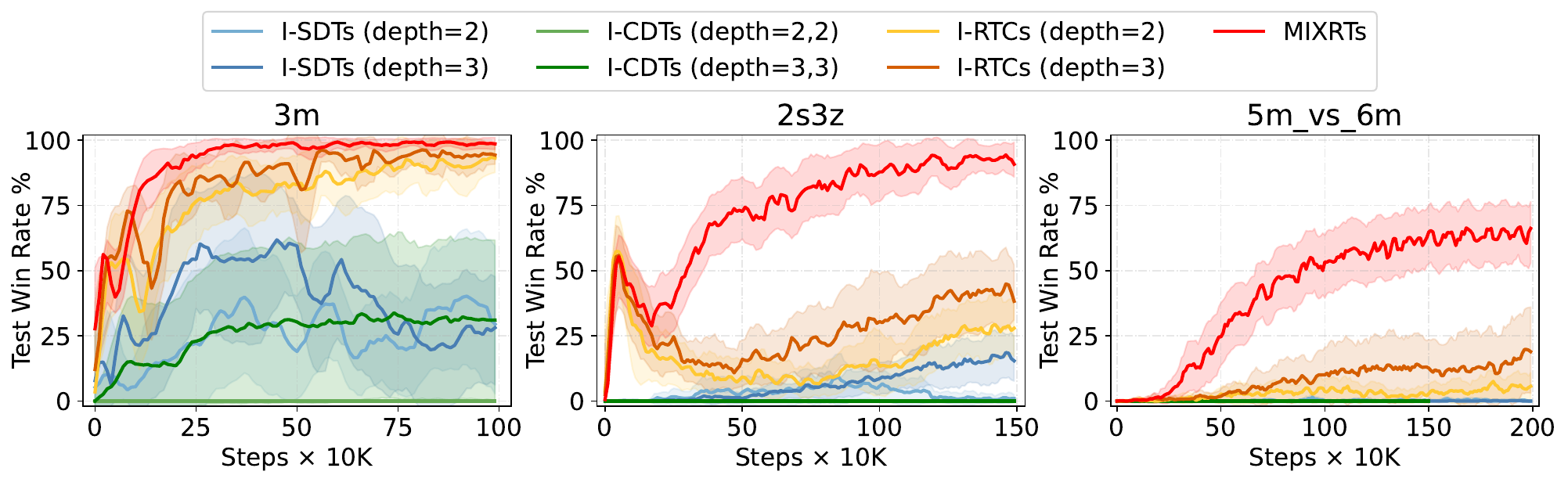}
	\caption{Comparison of I-SDTs, I-CDTs, I-RTCs, and MIXRTs with different depths.}
	\label{fig15}
\end{figure*}

During the training phase, agents receive rewards for defeating all enemy units within a limited time per episode, and the target network is updated after every $200$ training episode. 
We pause the training process for every $5000$ training timesteps and test the win rate of each algorithm for $32$ episodes using greedy action selection in a decentralized execution setting. 
All results are averaged over $8$ runs with different random seeds and are displayed in the style of $\text{mean} \pm \text{std}$.
Our model runs from $1$ hour to $15$ hours per task with an NVIDIA RTX 3080TI GPU and an Intel i9-12900k CPU, depending on the complexity and the length of the episode in each scenario.
Our code is publicly available at \url{https://github.com/zichuan-liu/MIXRTs}.

\begin{figure*}[tb]
	\centering 
	\subfigbottomskip=0pt 
	\subfigcapskip=-0pt 
	\setlength{\abovecaptionskip}{5pt}
	\subfigure[Comparison of different depths on different scenarios.]{	\label{fig13}
		\includegraphics[width=.99\linewidth]{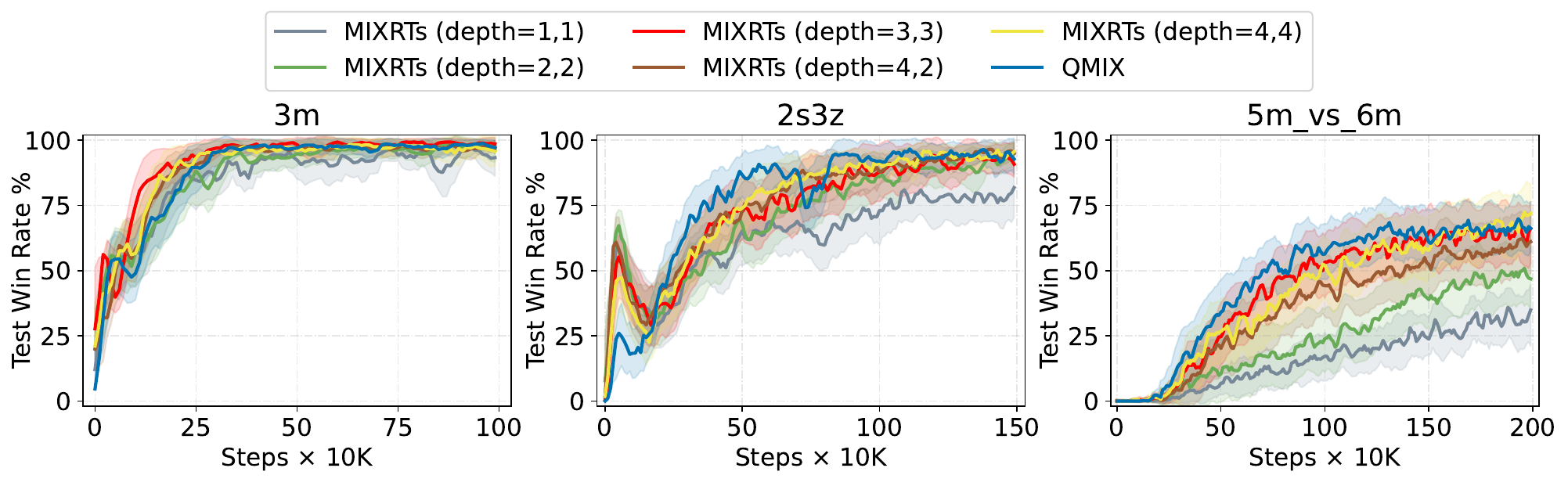}}	\\
	\subfigure[Comparison of different ensemble sizes on different scenarios.]{ 	\label{fig14}
		\includegraphics[width=.99\linewidth]{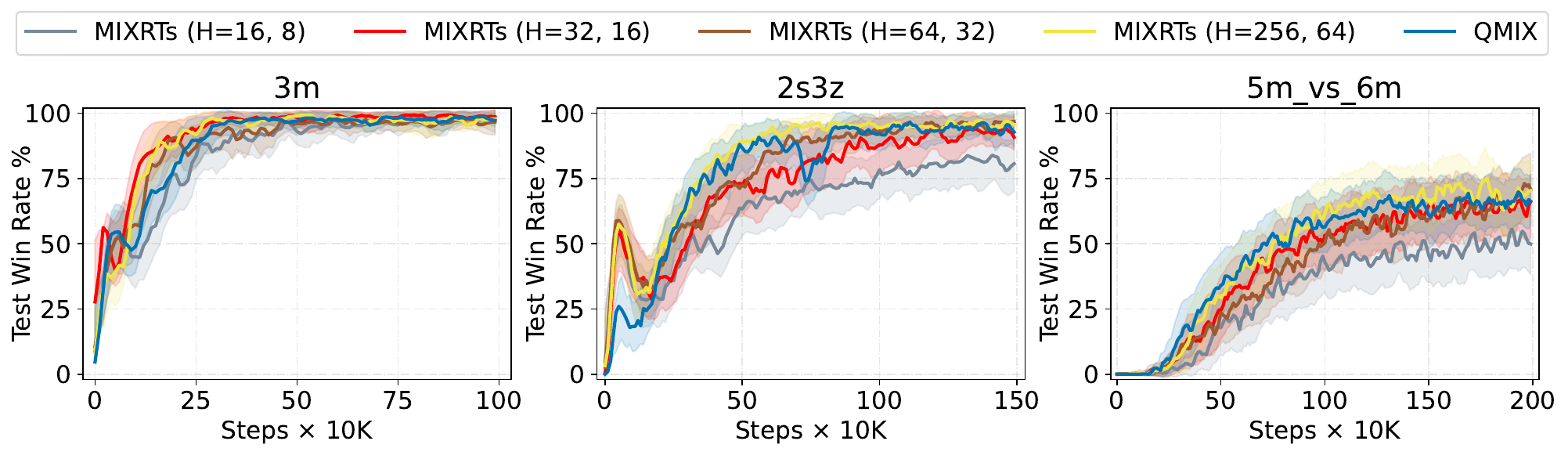}}
	\caption{The effect of the depth and the number of ensemble trees on MIXRTs performance.}
\end{figure*}

\begin{figure*}[tb]
	\centering 
		\subfigbottomskip=0pt 
		\subfigcapskip=-5pt 
	\subfigure[]{	\label{fig7a}
		\includegraphics[width=.79\linewidth]{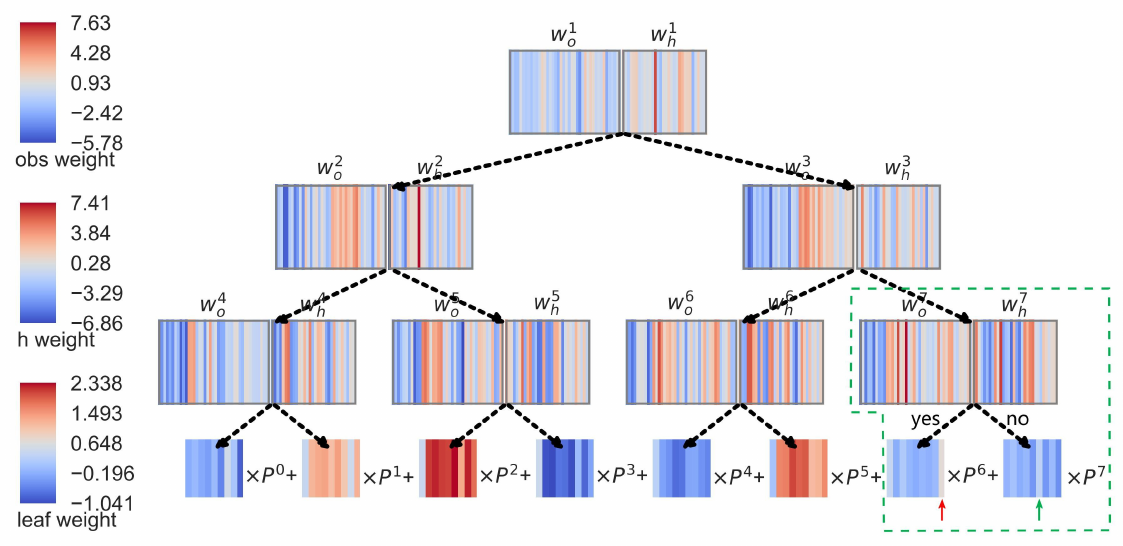}}	
	\subfigure[]{ \label{fig7b}
		\includegraphics[width=.18\linewidth]{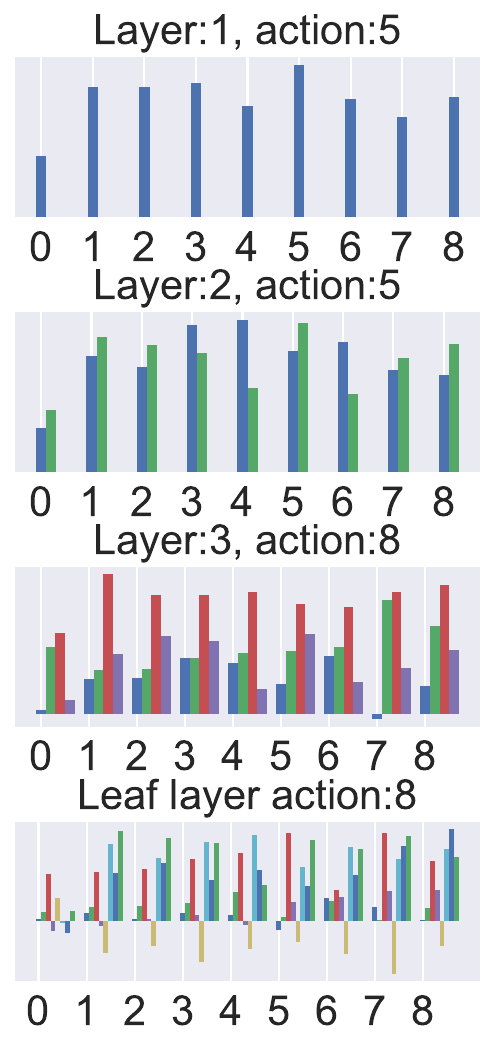}}
	\caption{Heatmap visualization of earned filters and action distributions for each layer.
		(a) Heatmap visualization of the learned filters in the learned RTCs of depth $3$. The weights of each non-leaf node feature contain the current observations (left) and the historical records (right), respectively. The leaf nodes indicate the magnitude of the different action distributions. (b) Actions probability distribution of the nodes of the RTCs with a given observation, where each node is distinguished by a different color bar.}
		\vskip -0.1in
\end{figure*}

\subsection{Performance Comparison}	
\textbf{Performance comparison in MPE.} We first conduct the experiments on three specifically tailored Spread tasks. 
As shown in Fig.~\ref{figfafasf15}, our method achieves competitive performance compared to baselines, demonstrating its efficiency across a range of scenarios.
In the standard task configuration involving $3$ agents, all algorithms except QPLEX successfully learn effective policies that result in covering an average of $95\%$ of the landmarks.
Notably, in scenarios with increased complexity featuring $4$ or $5$ agents, MIXRTs demonstrate superior performance in later iterations, achieving a consistent strategy that successfully covers $80\%$ to $90\%$ of the landmarks within $4M$ steps. 
In contrast, other algorithms display significant performance volatility.
This suggests that deploying a linear structural tree in policy learning is more conducive to consistently optimal strategy, i.e., each agent is tasked with occupying a distinct landmark.

\textbf{Performance comparison in SMAC.} 
First, we validate MIXRTs on a range of easy scenarios. 
As shown in Fig.~\ref{fig4}, compared to VDN, QMIX, QTRAN, and QPLEX methods, MIXRTs achieve competitive performance with a slightly faster learning process.
In the homogeneous scenarios (e.g., 3m and 8m), MIXRTs perform slightly better than others in the early learning stage.
The baselines obtain a sub-optimal strategy with a slightly larger variance in win rates compared to the MIXRTs.
QTRAN performs not well in these comparative experiments, which may suffer from the relaxations in practice impeding its precise updating~\cite{wang2020qplex}.
Especially in the heterogeneous 2s3z map, the MIXRTs still obtain a competitive performance whose win percentage is near to QMIX and QPLEX, but slightly higher than VDN and QTRAN, which may benefit from the efficient value decomposition via the lightweight inference.

Next, we evaluate the performance of different algorithms on the hard and super-hard scenarios.
As illustrated in Fig.~\ref{fig4}, MIXRTs consistently achieve performance close to the best baseline on different challenging scenarios and even exceed it on 6h\_vs\_8z.
Compared to the other baselines, MIXRTs perform slightly better in the three difficult scenarios, highlighting the model's ability to strike an optimal balance between interpretability and cooperative learning performance.
Moreover, we compare the performance of different mixing architectures in these scenarios, as shown in Table~\ref{sbreviewers}.
The results show that the factorization employed within the mixing trees yields a more effective computation of joint action values than VDN and QTRAN.
For super hard tasks 6h\_vs\_8z and MMM2, MIXRTs can search for workable strategies with stable updates compared to most baselines, indicating that our lightweight inference and ensemble structure can improve the learning efficiency and stability on the asymmetric scenarios.
To summarize, MIXRTs achieve competitive performance and stable learning behavior while retaining an interpretable learning architecture.

\begin{table}[tb]
	\caption{Test win rates \% of different mixing architectures in hard scenarios, where the methods are named individual function + mixing networks.
 }\label{sbreviewers}
	\centering
     \resizebox{1\columnwidth}{!}{
	\begin{tabular}{l|cccc}
		\hline
		\hline
		Method & 5m\_vs\_6m & 3s5z & 8m\_vs\_9m \\
		\hline
		RNN + VDN Net& ${73.95}_{\pm8.41}$ & ${82.03}_{\pm8.11}$ & ${75.78}_{\pm14.01}$\\
  		RNN + QTRAN Net& ${55.46}_{\pm11.26}$ & ${17.96}_{\pm14.31}$ & ${71.09}_{\pm17.18}$\\
  	RNN + Mixing Trees & ${69.17}_{\pm15.38}$ &${87.89}_{\pm5.04}$ & ${78.90}_{\pm10.39}$\\
  	RTCs + Mixing Trees & ${71.87}_{\pm12.73}$ &${83.98}_{\pm11.79}$ & ${84.76}_{\pm9.16}$\\
		\hline
		\hline
	\end{tabular}
 }
\end{table}

In addition, we also analyze the simplicity of MIXRTs in terms of the number of parameters compared to the above baselines, which depends mainly on the observation size and the number of agents on different scenarios.
As shown in Table~\ref{table1}, VDN requires fewer learnable parameters than QMIX, QTRAN, and QPLEX because it does not employ a mixing network to represent the state-value function.
Since each layer of MIXRTs is represented linearly, the number of parameters increases linearly with the depth of the tree.
When the depth of the RTCs is set to $3$, compared to QMIX, QTRAN, and QPLEX, the number of parameters of MIXRTs is reduced by more than $49\%$, whereas it retains competitive performance.
The simplicity of the model allows us to more easily understand the critical features and the decision-making process through lightweight inference, which strikes a better balance between performance and interpretability.

\textbf{Performance comparison with representative interpretable models based on the tree structure.}\label{DifferentTreePerformance}
We evaluate the learning performance of RTCs over the individual SDTs and CDTs methods on both easy and hard scenarios.
As shown in Fig.~\ref{fig15}, I-RTCs constantly and significantly outperform I-SDTs and I-CDTs in performance and stability across all the tasks.
I-RTCs achieve better performance by combining high-dimensional features mapped from observations to historical information as input, where adding recurrency information can help agents better capture feature information in complex tasks, especially in non-stationary multi-agent tasks.
In terms of stability, I-RTCs perform more stably than SDTs and CDTs, owing to its ensemble mechanism.
Compared to the existing interpretable tree-based methods, MIXRTs outperform I-SDTs, I-CDTs, and I-RTCs by leveraging the advantages of learning a centralized but factorized joint action-observation value, particularly through their lightweight inference architecture.

Generally, deeper trees tend to have more parameters, which can compromise interpretability.
Here, we further analyze the stability of the decision tree methods with different depths across different scenarios.
As shown in Fig.~\ref{fig15}, I-SDTs, I-CDTs, and I-RTCs learn faster and perform better as the depth of the tree increases.
Notably, by utilizing the ensemble tree structure and advanced recurrent techniques in I-RTCs, we achieve better performance that is generally less sensitive to tree depth compared to I-SDTs and I-CDTs.
From the comparisons, MIXRTs yield substantially better results than other tree-based methods. 
This improvement may be attributed to the ability of MIXRTs to efficiently approximate the complex relationship between individual action-values $Q_i$ and the joint action-value $Q_{tot}$, as well as to capture different features in subspaces.

\begin{figure*}[!htb]
	\centering 
	\subfigbottomskip=0pt 
	\subfigcapskip=0pt 
	\subfigure[Comparison of feature importance for MIXRTs on 8m.]{\label{figs3a}
		\includegraphics[width=.75\linewidth]{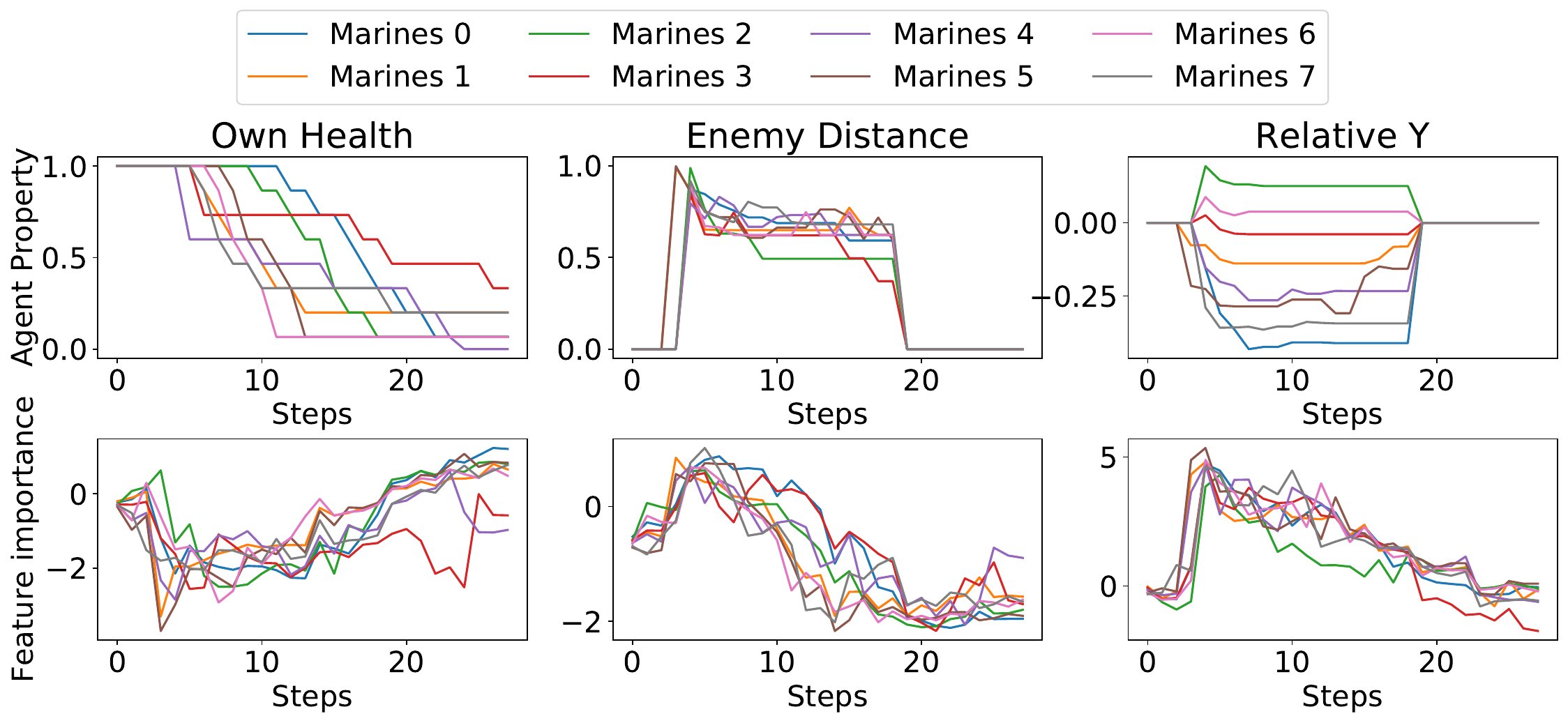}}
	\subfigure[Agent weight heatmap on 8m.]{ \label{figs3b}
		\includegraphics[width=.16\linewidth]{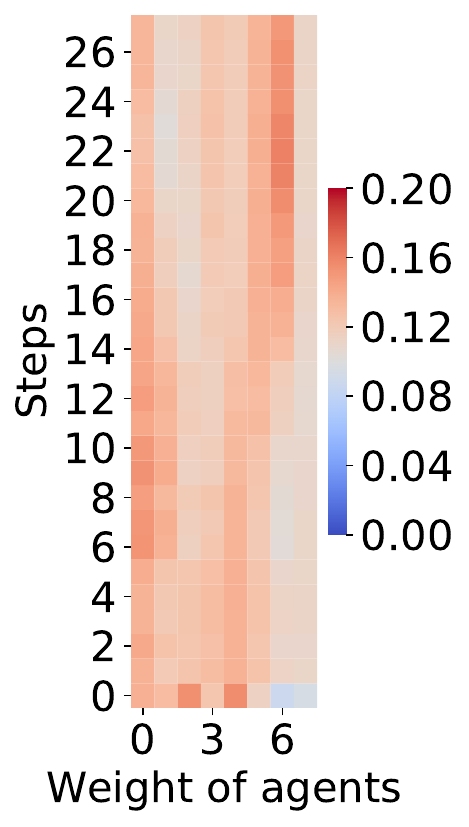}}\\
	\subfigure[Comparison of feature importance for MIXRTs on 2s3z.]{\label{figs3c}
		\includegraphics[width=.75\linewidth]{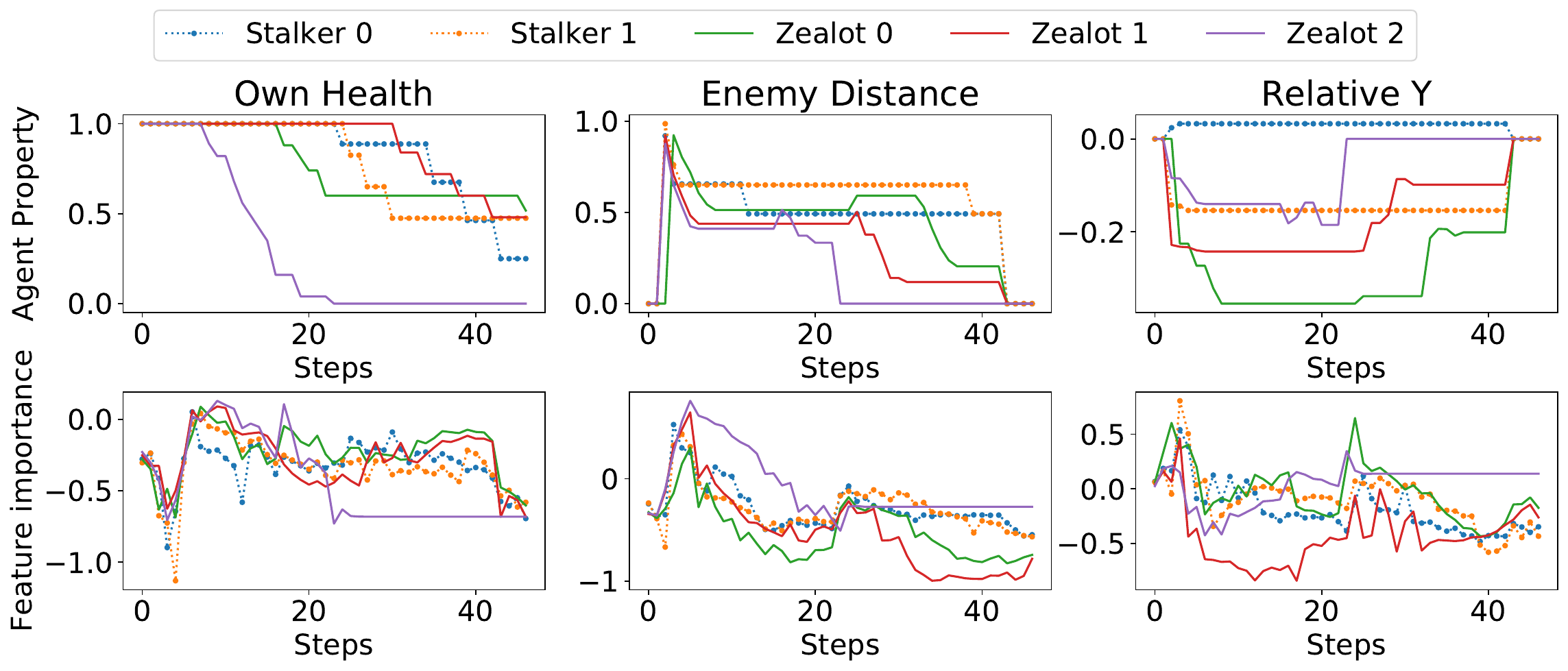}}
	\subfigure[Agent weight heatmap on 2s3z.]{ \label{figs3d}
		\includegraphics[width=.16\linewidth]{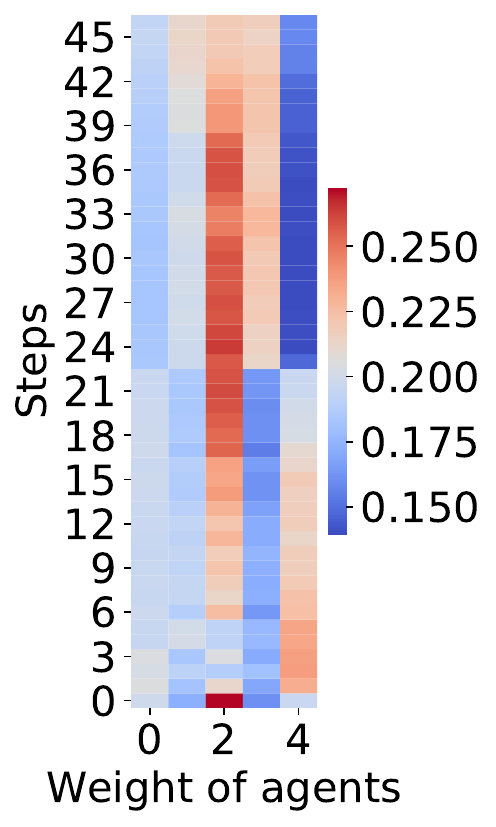}}\\
	\subfigure[Comparison of feature importance for MIXRTs on MMM2.]{\label{figs3e}
		\includegraphics[width=.75\linewidth]{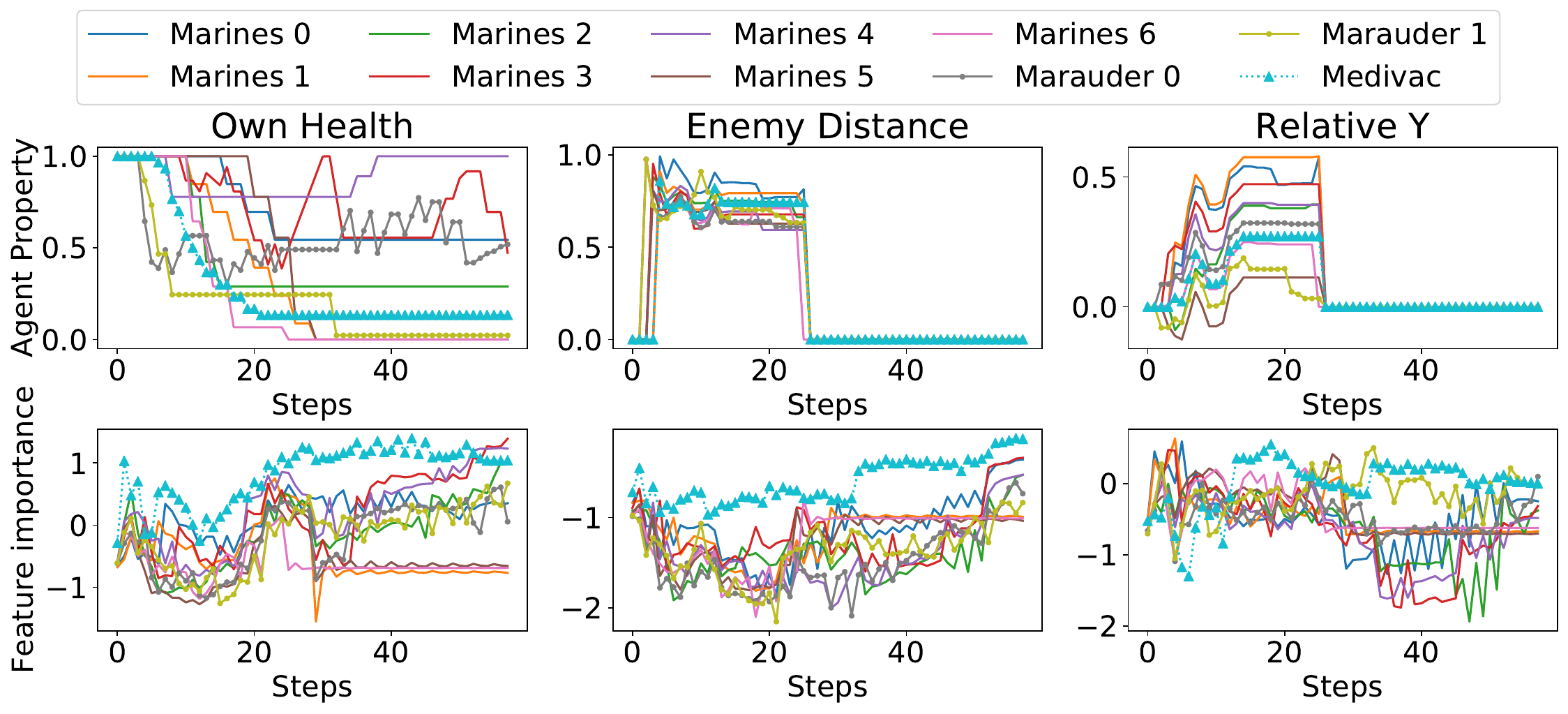}}
	\subfigure[Agent weight heatmap on MMM2]{ \label{figs3f}
		\includegraphics[width=.16\linewidth]{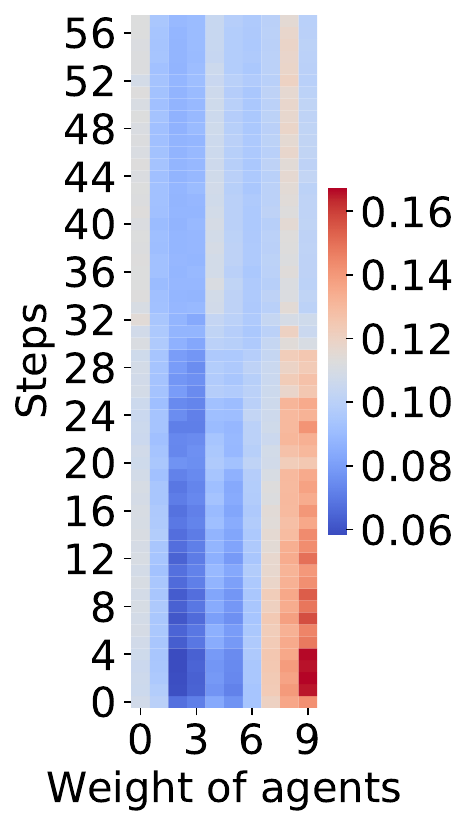}}\\
	\caption{Feature importance and assigned weights of $Q$-values. 
    As the number of steps increases, feature importance within the line graph varies correlating to the properties of relevant agents.
    In the heatmaps, horizontal and vertical ordination indicate the agent-id and steps, respectively.
 }
 \vskip -0.1in
 \label{fig8}
\end{figure*}

\textbf{Sensetivity analysis.} \label{ablation}
In addition, we analyze the sensitivity of MIXRTs to the effects of tree depth and the number of ensemble trees on performance.
Here, we vary these factors to assess how performance is influenced by the tree depth of the individual RTCs and the mixing trees of MIXRTs.
Fig.~\ref{fig13} and Fig.~\ref{fig14} show the respective effects of the above factors on the performance of MIXRTs.
First, we investigate the influence of the individual RTCs and the mixing trees with different depths in three scenarios.
Fig.~\ref{fig13} displays that the performance of MIXRTs will improve as the depth of the individual RTCs and the mixing of RTCs appropriately increases.
A greater depth allows MIXRTs to produce more fine-grained behaviors, leading to better model performance.
However, the improved performance comes at the trade-off of losing its interpretability.
Generally, a moderate depth (e.g., depth=3, 3) setting can obtain a competitive performance, where the former $3$ and the latter $3$ represent the depth of the individual action-value RTCs and mixing RTCs of MIXRTs with depth=$3, 3$, respectively.
Further, we study the effect of the number of ensemble trees on performance.
Fig.~\ref{fig14} indicates that the tree model tends to be unstable when the number of ensemble trees is small. 
With moderate values (e.g., $H=32 \sim 64$), the tree usually converges quickly and obtains better performance.
This suggests that setting a moderate tree depth and number of ensemble trees can yield promising performance while retaining the simplicity and interpretability of the model. 
For this reason, we chose moderate parameters that offer a trade-off between performance and interpretability, while maintaining a lower number of parameters comparable to the baselines, as shown in Table~\ref{table1}.

\section{Interpretability}\label{InterpretabilityInterpretability}
The main motivation for this work is to create a model whose behavior is easy to understand, mainly by fully understanding the decision process along the root-to-leaf path and their roles in the team.
To demonstrate the interpretability provided by our method, we describe the structure of RTCs through the learned filters at inner nodes and show the visualization of the learned action distribution.
Furthermore, we present the importance of input features, describe how they influence decision-making, and explore the stability of feature importance.
Finally, we give user studies to ensure that the interpretations align with human intuition.

\subsection{Explaining Tree Structure.}\label{Interpretability}
The essence of RTCs is that a kind of model relies on hierarchical decisions rather than hierarchical features.
The neural network generally allows the hierarchical features to learn robust and novel representations of the input space, but it will become difficult to interpret once more than one level.
In contrast, we can immediately engage with each decision made at a higher level of abstraction, where each branching node of the RTCs directly processes the entire input features.
By filtering different weights to each feature at each branching node, it becomes possible to understand which features the RTCs consider when assigning a particular action distribution to a specific state and how these features influence the selected actions.
This understanding is achieved by simply examining all the learned filters along the traversed path from the root to the leaf.

As shown in Fig.~\ref{fig7a}, we display the structure of the learned RTCs model with a depth of $3$ for each agent on a 3m map, where the arrows and lines indicate the connections among tree nodes.
Each node assigns different weights to each feature by processing the observed state, where a feature with a more intense color (positive: red, negative: blue) indicates higher magnitude weights and receives more focus. 
Taking the node in the green dotted area as an example, the three features with more intense color are at position $17$ (representing the relationship with the enemy of $\text{id}=2$), at position $14$ (representing whether this enemy is visible or not), and at position $29$ (representing its health value).
Since these features with higher weights tend to direct the decision towards the left leaf, it becomes important to choose to attack this enemy in the action distribution (marked by a red arrow).
Otherwise, the leaf distributions assign probabilities to actions related to moving westward (marked by a green arrow).
This behavior is intuitive, as attacking an enemy can bring higher rewards when it has a better chance of survival.
Therefore, depending on the choice generated by the activation function, the selected nodes focus on the values of different branching decisions, including action and feature weights.

To understand how a specific state observation influences a particular action distribution, we visualize the decision route from the root to the chosen leaf node in RTCs with the input state.
For a given observation, we also provide the action probability distribution at each layer, which is an inherent interpretation capability not available by the standard DNNs paradigm.
As shown in Fig.~\ref{fig7b}, we can receive the action probability distribution from each layer.
Each node outputs a probability distribution associated with a feature vector, and the selected action depends on the probabilities obtained by linearly combining the action distributions of each leaf node.
From Fig.~\ref{fig7b}, we can find which features obtain more attention at each layer and how these features affect the action probability distribution.

\begin{table*}[tb]
	\caption{Perturbation tests the win rate of MIXRTs. To quantify the interpretation, we use the trained model to calculate the important features in each step via Eq.~(\ref{Eq14}).
Then, we mask a varying percentage of the least and most important features with zeros for each step and redo the decision-making.  }\label{Perturbation}
	\centering
	\fontsize{9}{12}\selectfont 
	\begin{tabular}{l|ccc|l|ccc}
		\hline
		\hline
		Least important data & 3m& 2s3z & 5m\_vs\_6m & Most important data & 3m& 2s3z & 5m\_vs\_6m\\
		\hline
		Masking  0\%  & ${100.00}_{\pm0.00}$ & ${100.00}_{\pm0.00}$ & ${63.37}_{\pm2.10}$ & 
  Masking  0\% &  ${100.00}_{\pm0.00}$ & ${100.00}_{\pm0.00}$ & ${63.37}_{\pm2.10}$ \\
  	Masking 5\% & ${100.00}_{\pm0.00}$ &${100.00}_{\pm0.00}$ & ${48.95}_{\pm1.72}$&
      Masking 5\% & ${83.48}_{\pm1.35}$ & ${93.87}_{\pm1.18}$ & ${13.37}_{\pm3.78}$\\
		Masking 10\%  & ${96.77}_{\pm1.74}$ & ${100.00}_{\pm0.00}$ & ${41.75}_{\pm2.57}$&
  		Masking 10\%  & ${63.78}_{\pm3.89}$ & ${41.19}_{\pm0.73}$ & ${0.00}_{\pm0.00}$
  \\
		Masking 20\%  & ${82.19}_{\pm1.46}$ & ${70.96}_{\pm5.26}$ & ${15.53}_{\pm3.24}$&
  Masking 20\%  & ${12.90}_{\pm1.51}$ & ${3.22}_{\pm2.64}$ & ${0.00}_{\pm0.00}$\\
		Masking 30\%  & ${48.38}_{\pm4.15}$ & ${46.67}_{\pm3.
  19}$ & ${0.00}_{\pm0.00}$&
  Masking 30\%  & ${0.00}_{\pm0.00}$ & ${0.00}_{\pm0.00}$ & ${0.00}_{\pm0.00}$\\
		\hline
		\hline
	\end{tabular}
\end{table*}

\subsection{Feature Importance}\label{dsafdaoh}
There are several ways of implementing feature importance assignments on SDTs~\cite{ding2020cdt}.
However, the data point is more susceptible to being perturbed since it is less confident of remaining in the original when there are multiple boundaries for partitioning the space.
To mitigate this issue, we utilize decision confidence as a weighting factor in feature importance assignment, which can be positively correlated with the distance from the instance to the decision boundary to relieve the above effects. 
Therefore, similar to Eq.~(\ref{Eq6}), we weight the confidence probability of reaching the deepest non-leaf level node $j$ as $P^j(o_{i}^t, h_i^{t-1})$. 
The feature importance can be expressed by combining these confidence values with the weights of each decision node
\begin{equation}\label{Eq14}
I(o_{i}^t) = \sum_{j}^{} P^j(o_{i}^t, h_i^{t-1}) w^j_o.
\end{equation}

\textbf{Feature importance analysis.} After obtaining learned RTCs, we evaluated the feature importance of agents on different maps with Eq.~(\ref{Eq14}). 
We select three agent feature properties to indicate their importance and display the weights of individual credit assignments across different episodes, including the agent’s health, enemy distance, and relative Y properties. 
We analyze these properties on three different maps, and the
results are shown in Figs.~\ref{figs3a},~\ref{figs3c}, and \ref{figs3e}, respectively. 
The horizontal coordinate represents the number of steps in the episode, and the two vertical coordinates represent the corresponding value of the agent property and feature importance, respectively. 
Meanwhile, we visualize the weights $W_i$ for each agent $i$ on each scenario, and the weight heatmaps are shown in Figs.~\ref{figs3b}, \ref{figs3d} and \ref{figs3f}.
In the attention heatmaps, the steps increase from bottom to top, and the horizontal ordination indicates the agent id.

For the 8m scenario, it is essential to achieve a victory that the agents avoid being killed and pay more attention to firepower for killing the enemies. 
From Fig.~\ref{figs3a}, we observe that allies agents have similar feature importance on the same attributes.
It is worth noting that when an enemy is killed in combat, both the importance of the distance to this enemy and its relative Y will decrease.
This indicates that RTCs can capture the skills for accomplishing tasks and generating coping actions in response to environmental changes.
In addition, as shown in Fig.~\ref{figs3b}, we can find that each agent has almost equal attention weights, indicating that each allied Marine plays a similar role in the homogeneous scenario, and MIXRTs almost equally divides $Q_{tot}$ for each agent. 
MIXRTs perform well with the refined mixing trees since they adjust the weights of $Q_i$ with a minor difference.
For the 2s3z scenario, we observe that Stalkers and Zealots have similar feature importance at different steps, which might be sourced from the fact that they all focus on their health and play an equally important role in the battle. 
Furthermore, when the health value of the Zealot $2$ drops to $0$, we find that it exhibits strongly negative importance with blue shades as shown in Fig.~\ref{figs3d}, where the agent does not play an active role in battle since it is killed. 
Interestingly, on the 2s3z scenario, there is a significant shift in the weight assignments among the agents. 
Following the elimination of Zealot $2$ during the battle and its subsequent replacement by Zealot $0$, we observe a marked increase in the importance of the distance and health features.
Moreover, we observe that different types of soldiers have different sensitivity to features. 
For example, Medivac (agent $9$ in Fig.~\ref{figs3f}) receives more attention during the early stages of the battle, which may be attributed to its unique role as a support unit in combat.
In summary, by analyzing the feature's importance, we receive meaningful implicit knowledge about the tasks, which can facilitate human understanding of why a particular feature's importance leads to a specific action.

\textbf{Perturbing important features.}
Due to the large number of features involved in the decision-making process, we mask certain features by degradation of performance to measure their importance.
For trained MIXRTs, we first calculate the feature importance at each step through the nodes.
Then, we evaluate performance by substituting the most and least important features with zeros, implying that the corresponding nodes in the tree structure are not involved in the calculation.
The performance results of the win rate are presented in Table~\ref{Perturbation}.
When perturbing a portion of unimportant features, the performance remains relatively stable, which suggests that the critical decision-making process does not heavily rely on the full set of features.
On the contrary, perturbing the features consider red important leads to a significant decrease in performance, affirming the importance of these features in the decision-making process.
Furthermore, as the proportion of masked features increases, the win rate declines precipitously to zero when critical features are perturbed, whereas it does not when unimportant features are masked.
This demonstrates the effectiveness of MIXRTs in evaluating the importance of features and allows for excluding a portion of features that are inconsequential to the decision-making process.

\subsection{Stability on Feature Importance}
\begin{figure}[tb]
	\centering 
	\subfigbottomskip=0pt 
	\subfigcapskip=-0pt 
		\subfigure[The violin plot of feature importance on 3m.]{\label{stab1}
			\includegraphics[width=.85\linewidth]{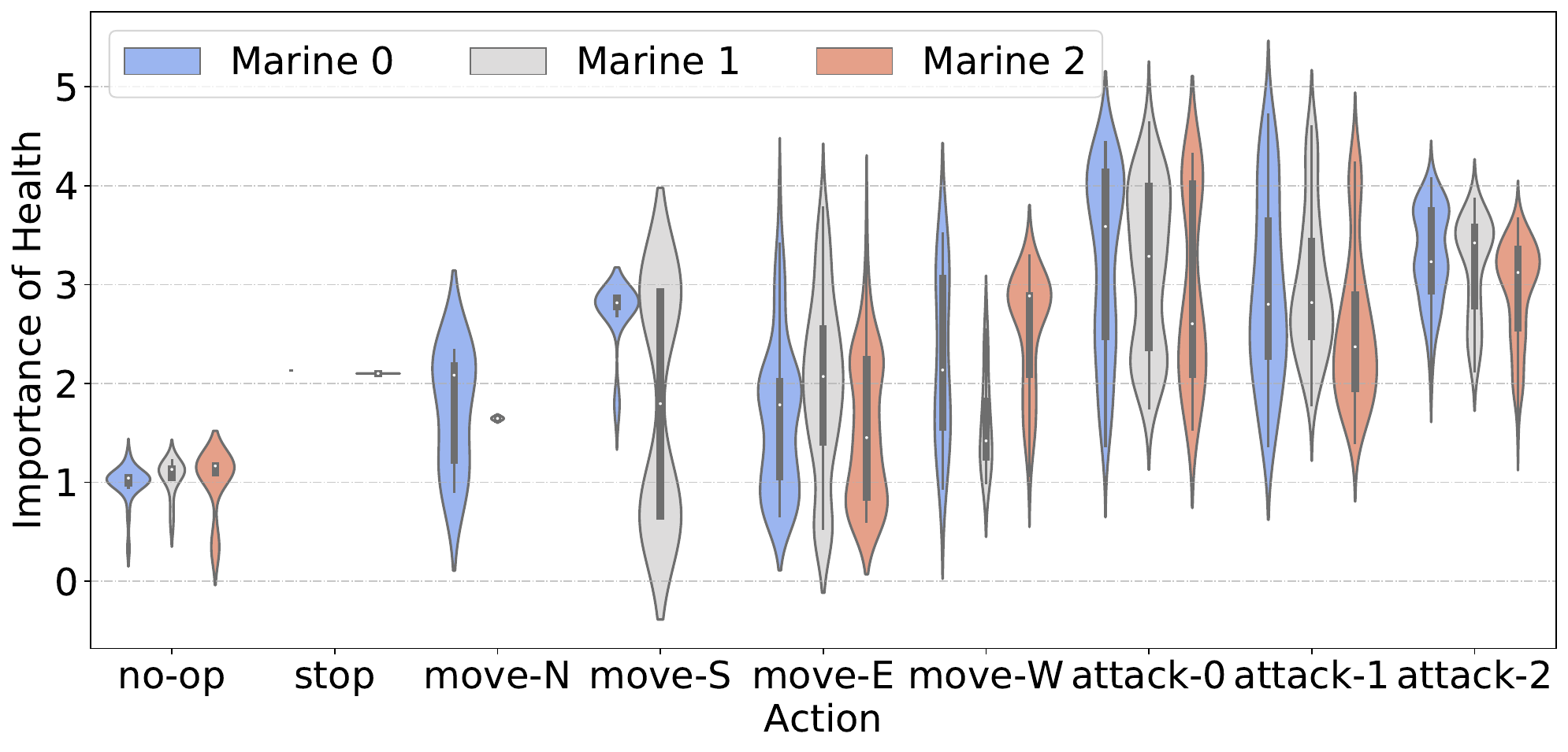}}\\
	\subfigure[The violin plot of  feature importance on 2s3z.]{\label{stab2}
		\includegraphics[width=\linewidth]{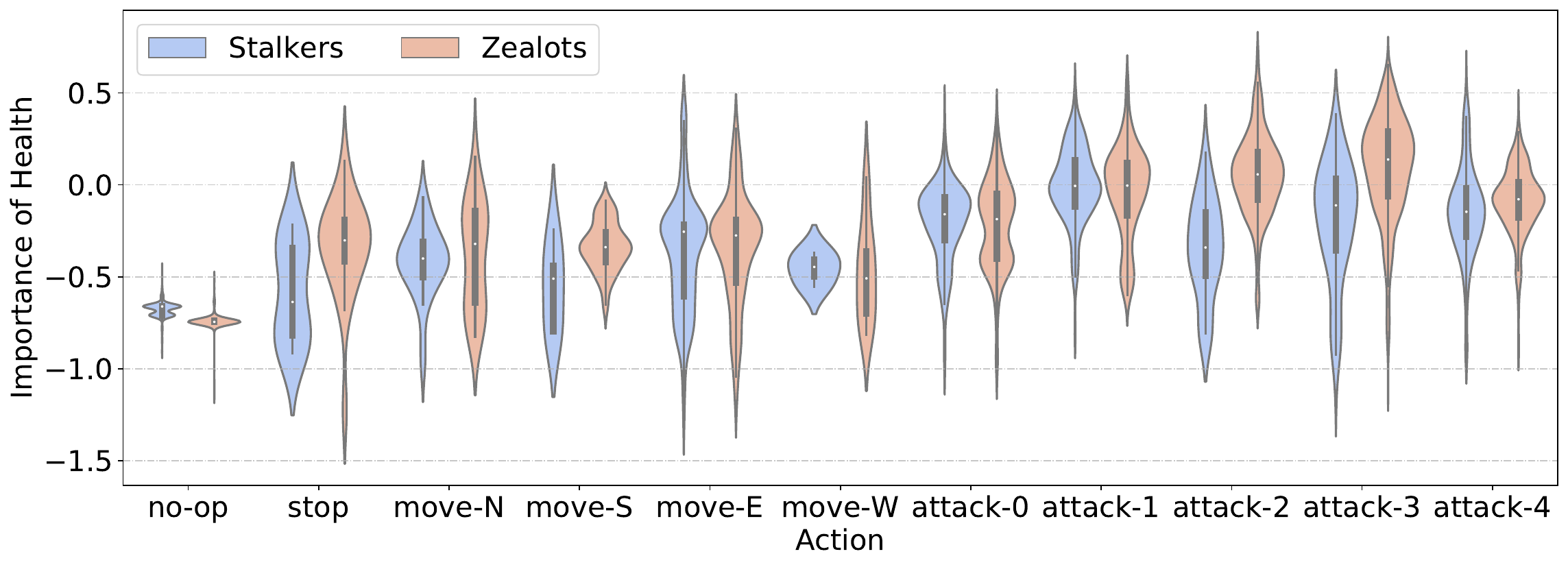}}\\
	\subfigure[The violin plot of  feature importance on MMM2.]{\label{stab3}
		\includegraphics[width=\linewidth]{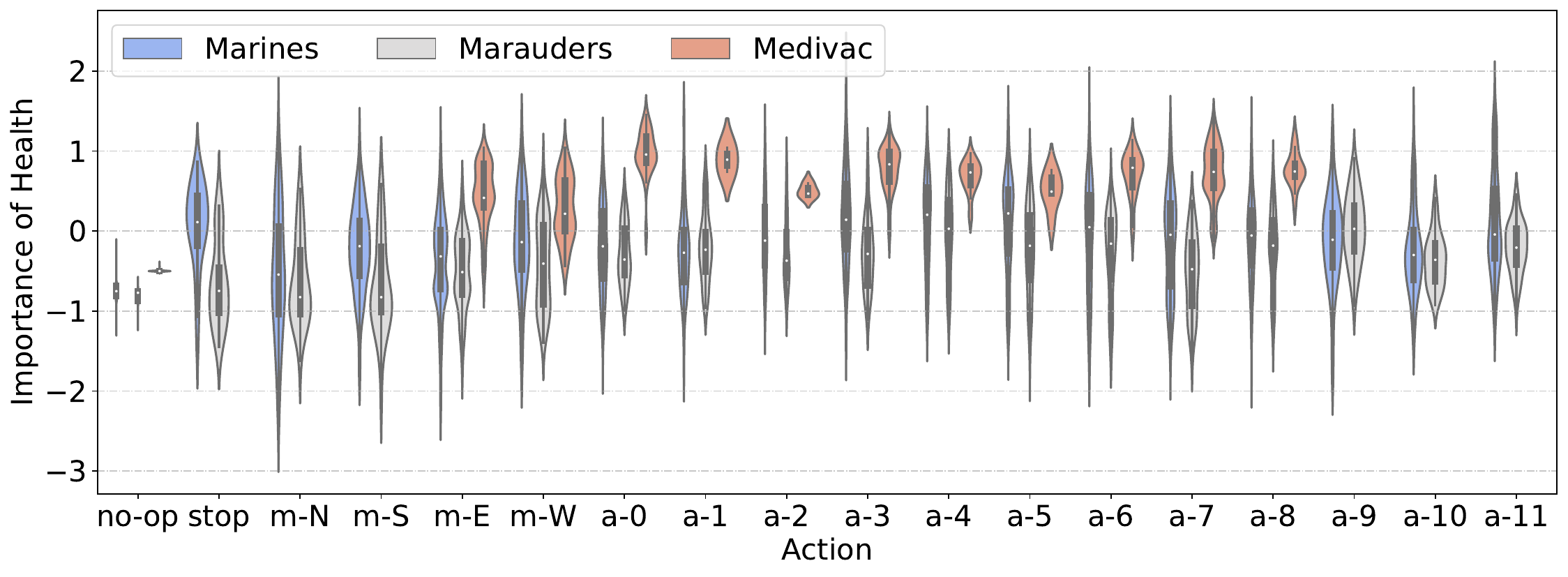}}\\
	\caption{Depicting the stability of fitted MIXRTs through violin plots.  Broad sections of the plot denote a higher density distribution, while narrow portions imply a less dense distribution.
}\label{stab3m}

\end{figure}
The stability of an interpretable model is an important factor reflecting reliability.
To further investigate stability, we delved into the feature importance assignments and action distributions of the MIXRTs.
We conduct analyses with the violin plot to study action distribution over several episodes.
The violin plot helps us to immediately identify the median feature importance without the need to visually estimate it by integrating the density, thereby providing more precise information to analyze the stability of feature importance.
We hope this can alleviate the differences in action distributions caused by the different initial states of each episode.
As shown in Fig.~\ref{stab3m}, we select the health property of the agents to indicate the correlation between the underlying action distribution and the assigned feature importance over $32$ episodes.
The horizontal coordinate and vertical coordinate represent the selected action and the importance of agent health, respectively.

On the 3m scenario, three allied agents focus on firepower to kill the enemies with fewer casualties.
As shown in Fig.~\ref{stab3m}, the agents exhibit similar importance across all actions, which aligns with the common knowledge that homogeneous agents play the same important role during the battles.
Besides, we find that they prioritize selecting attack actions, as these yield more positive rewards to ensure victory in this easy homogeneous environment.
From Figs.~\ref{stab2} and \ref{stab3}, considerable differences can also be spotted over $32$ episodes on the heterogeneous scenarios 2s3z and MMM2, even though MIXRTs have captured similar skills to win.
For the 2s3z scenario, Zealots display a higher importance of health than Stalkers, which may require a better winning strategy where Zealots agents serve at the front of combat, killing enemies one after another while protecting the Stalkers to kite the enemy around the map.
Furthermore, we notice that similar interesting findings also exist in the MMM2 scenario.
For example, Medivac, serving as a healer unit, receives more attention than other kinds of agents during the battle.
This focus on Medivac likely enhances the team’s effectiveness, as it uses healing actions to improve the health of its allies, which is consistent with the analysis in Section~\ref{dsafdaoh}.

Overall, the importance of health is slightly higher when the agents are in attacking status than in moving status, and it is significantly higher than when not operating.
Since agents are often being attacked while they are attacking, the importance of health is more intense.
Meanwhile, agents who are killed receive the negative importance of health with a small interquartile range.
Regarding stability, agents that are not operating maintain a steady state in the model, as their health value is $0$.
In summary, even when initial environmental conditions vary, we can still find implicit knowledge from MIXRTs by analyzing the feature importance over several episodes.

\subsection{Case Study}

\begin{figure}[tb]
	\centering 
	\subfigure[2s3z (step=10)]{\label{fig_adda}
		\includegraphics[width=.481\linewidth]{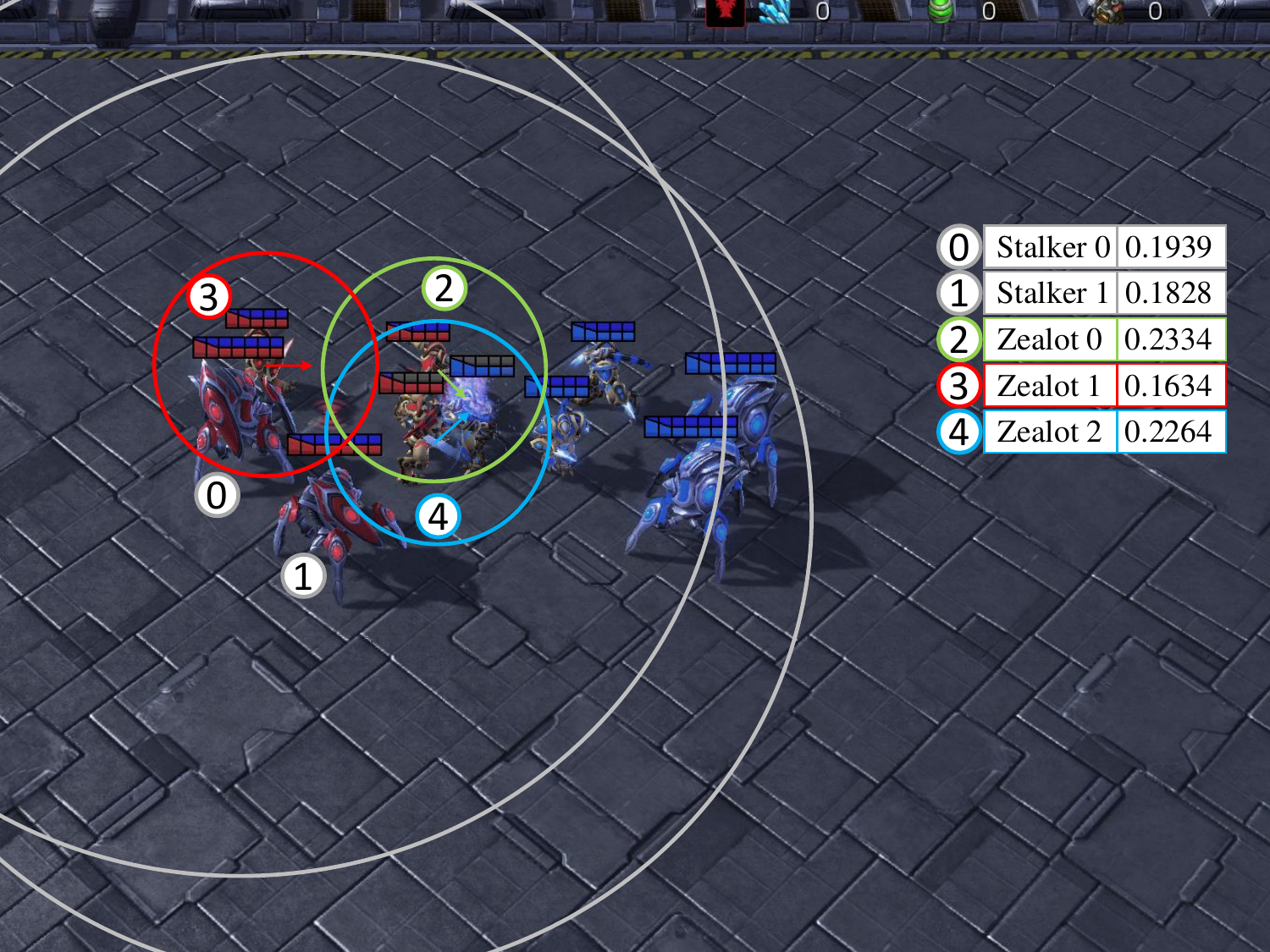}}
	\subfigure[2s3z (step=27)]{ \label{fig_addb}
		\includegraphics[width=.481\linewidth]{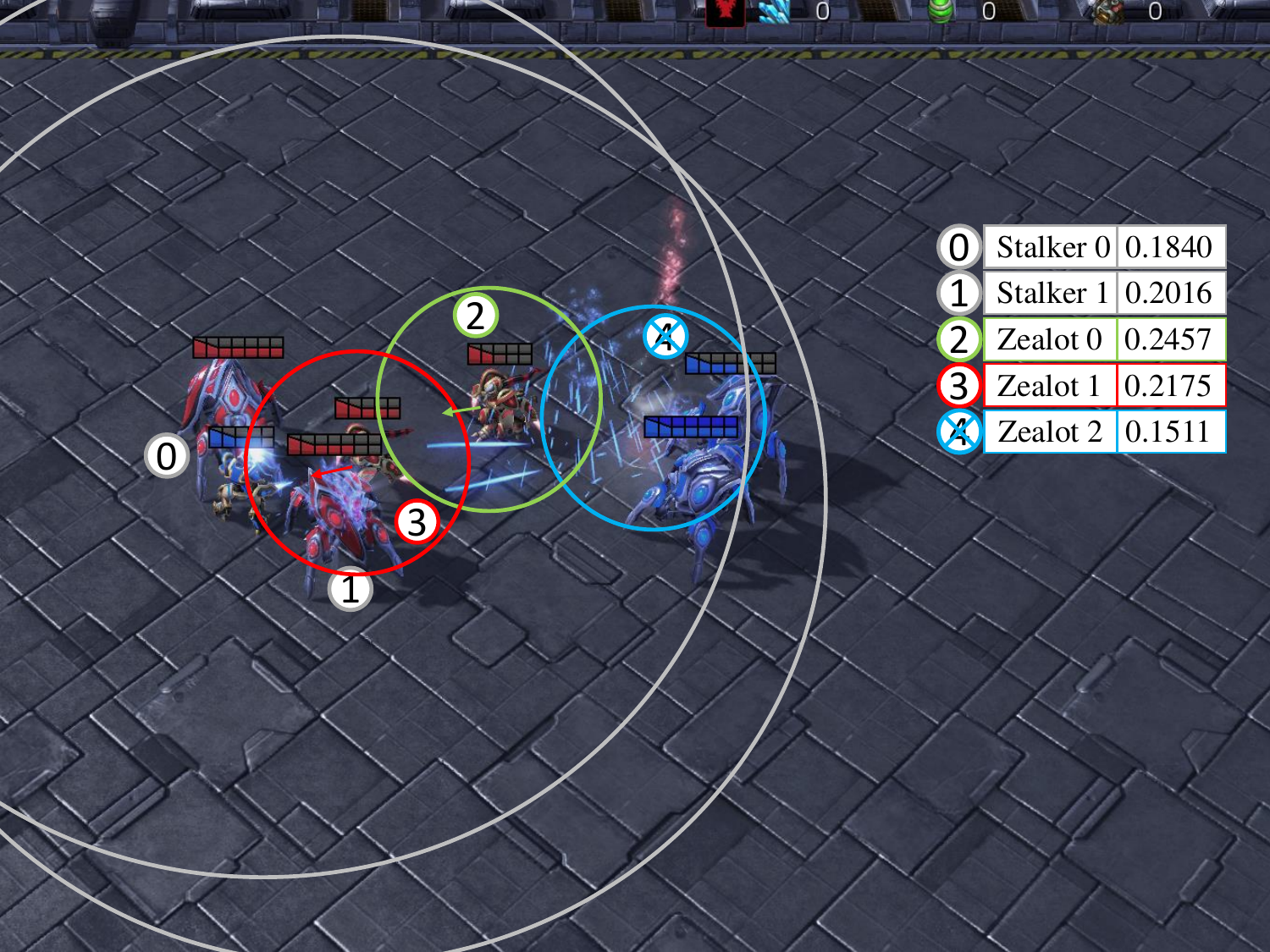}}
	\subfigure[2s3z (step=34)]{ 
		\includegraphics[width=.481\linewidth]{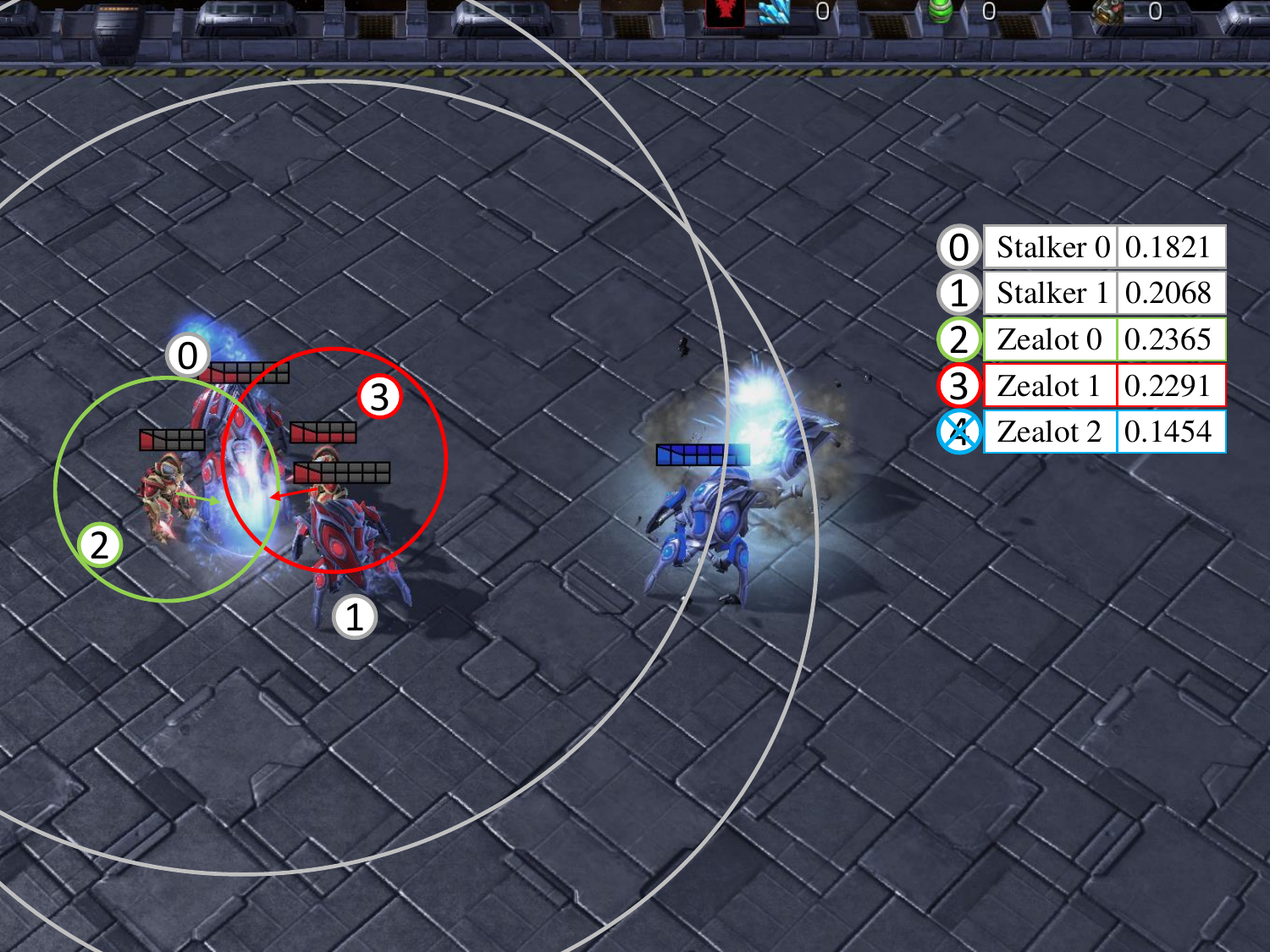}}
	\subfigure[MMM2 (step=7)]{ \label{fig_addd}
		\includegraphics[width=.481\linewidth]{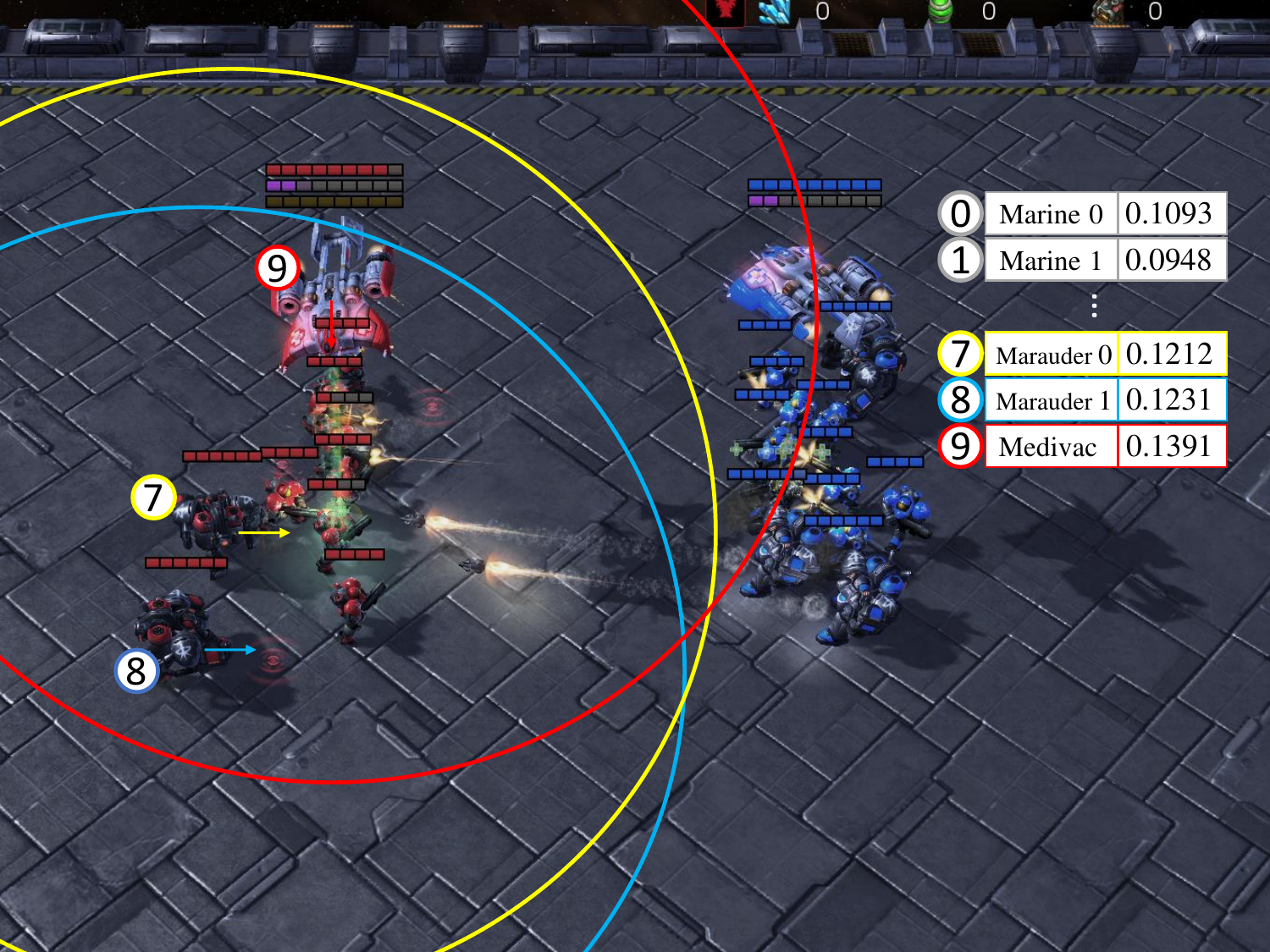}}
	\subfigure[MMM2 (step=12)]{ 
		\includegraphics[width=.481\linewidth]{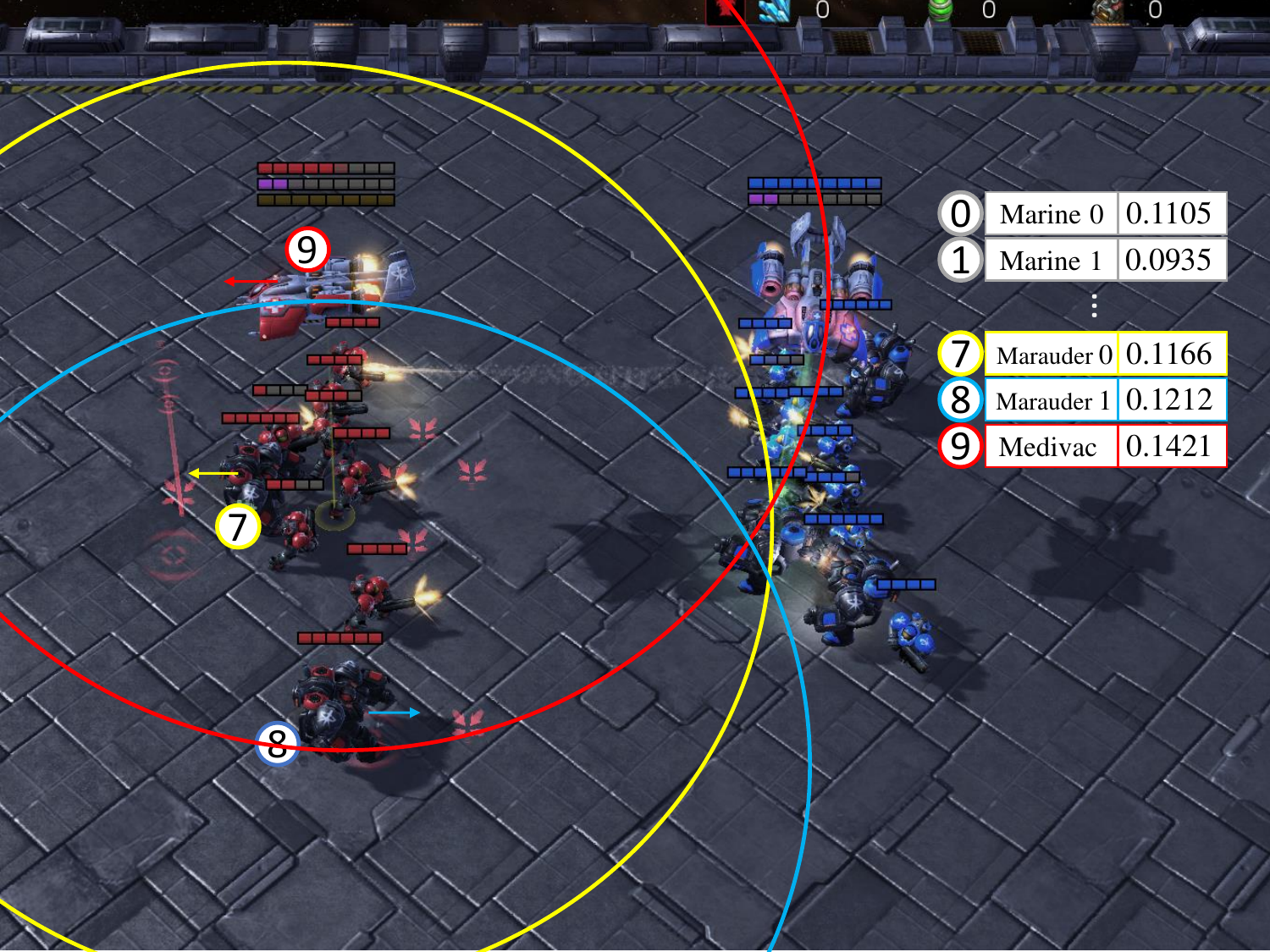}}
	\subfigure[MMM2 (step=35)]{ \label{fig_addf}
		\includegraphics[width=.481\linewidth]{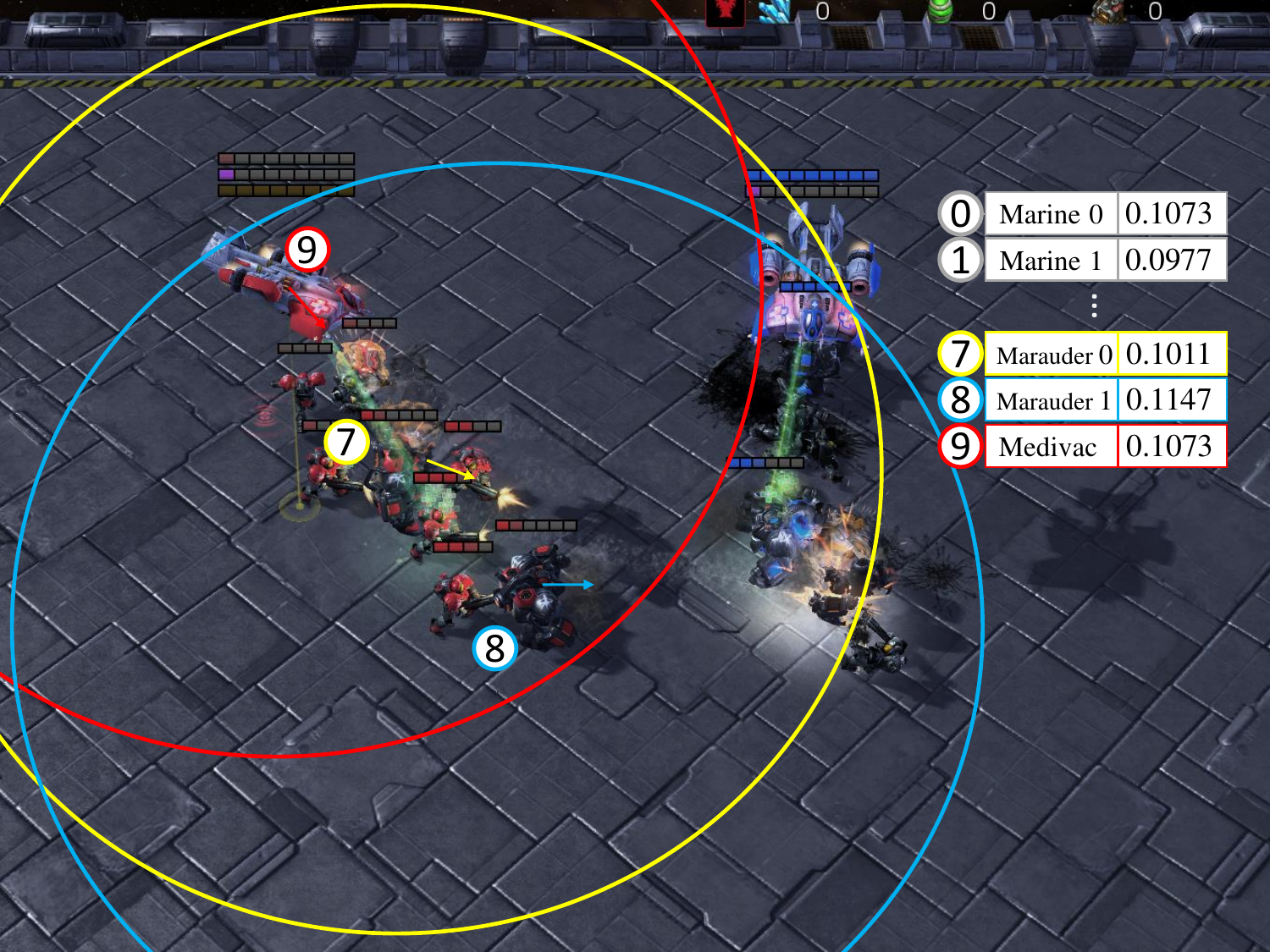}}
	\caption{
	Visualization of the evaluation for MIXRTs on 2s3z and MMM2 scenarios in SMAC (time step goes from left to right). 
	For each task, the credit assignment weights for each agent are displayed at the top-right of each frame. 
	The arrow and the colored circle indicate the direction that each moving agent is facing and the central attack range, respectively.
	For the 2s3z scenario, the agents will receive higher credit when they focus on killing the enemy within its range of attacking, which implies the credits reflect their contributions to the team.
 	As shown in figures (a) and (b), they learn a skill-winning strategy of blocking enemy attacks launched from different directions.
	For the MMM2 scenario, the assigned credit for each unit indicates the important role that Medivac receives the highest credit during the early stage of the battle since it can heal the allied units, and Marauders receive higher credit than Marines for their high damage.
	}\label{fig_add}
\end{figure}

To further demonstrate the interpretability of MIXRTs, we present the keyframes on these episodes during the testing stage in Fig.~\ref{fig_add}.
As observed from Fig.~\ref{fig_adda}, during the initial stage (step=$10$) of the 2s3z scenario, Zealot~$2$ and Zealot~$0$ are actively participating in the battle, while Zealot~$1$ drops teammates.
It can be understood that Zealot~$2$ and Zealot~$0$ make contributions to the team, whereas Zealot~$1$ is not significantly involved.
Their credit assignment weights $0.23344$, $0.22639$, and $0.16338$ correctly catch the manner of Zealot~$0$, Zealot~$2$ and Zealot~$1$, respectively.
When Zealot~$2$ was killed as shown in Fig.~\ref{fig_addb}, its credit assignment weight also quickly dropped to the lowest team level, while the weights of its allies increased.
During the battle process, both Stalkers capture high-level skills with stable credit assignment weights when the team is in a combat tactic, where they nearly always attack enemies behind teammates to get the utmost out of their long-range attack attributes.
Besides, Zealot~$0$ and Zealot~$1$ focus on killing the enemy, obtaining higher credit assignment weights at step $34$.
As a result, Zealot~$0$ and Zealot~$1$ contribute more than the others with greater credits, indicating that the credit assignments accurately reflect their contributions to the team.
For the MMM2 scenario, we find an interesting strategy to win the battle: when the Medivac is under attack, it retreats out of the enemy’s range while its allies advance to shield it from further attacks.
Its credit assignment weight correctly catches the important role of its team with the highest credit during the early stage of the battle. 
Besides, Marauders (Agents $7$ and $8$) receive higher credits due to their significant damage output, they receive healing blood from the Medivac agent as shown in Figs.~\ref{fig_addf}.
In summary, MIXRTs can help us understand the behaviors of agents and their contributions to the team through linear credit assignments on different complex scenarios.

\subsection{User Study}
Our final evaluation investigates the effect of the interpretability provided by MIXRTs for non-expert users.
Following the methodology of Silva et al.~\cite{silva2020optimization, silva2021encoding}, we design a user study where participants are shown policies trained on SMAC scenarios.
They are tasked with identifying whether these agents' credit assignments are reasonable given a set of observation inputs and keyframes\footnote{Due to the overwhelmingly large features, we only interpret credits to alleviate frustration for non-expert participants} (similar to Fig. ~\ref{fig_add}).
We compare the credit assignment rationality among MIXRTs, VDN, and QMIX.
We present our results to $20$ participants, instructing each to assign a helpfulness score from $1$ to $5$ using a Likert scale to each result.
The Likert scale guideline is provided on each score for justified scoring as
\begin{itemize}
\item Score $1$: misleading credits that impair the decision-making process.
\item Score $2$: uninformative credits that neither aid nor hinder human understanding.
\item Score $3$: deviated but somewhat related credits that provide slight assistance.
\item Score $4$: strongly related keyframes but slightly inaccurate, helpful.
\item Score $5$: well-aligned credits that contain all information in combat, exceptionally helpful.
\end{itemize}
Each participant is required to observe the keyframes without knowing which method produced the policy, ensuring full fairness.
For each method, we sample $22$ frames and time them separately.

\begin{figure}[tb]
	\centering 
	\subfigbottomskip=0pt 
	\subfigcapskip=-5pt
	\setlength{\abovecaptionskip}{5pt}
	\includegraphics[width=.47\textwidth]{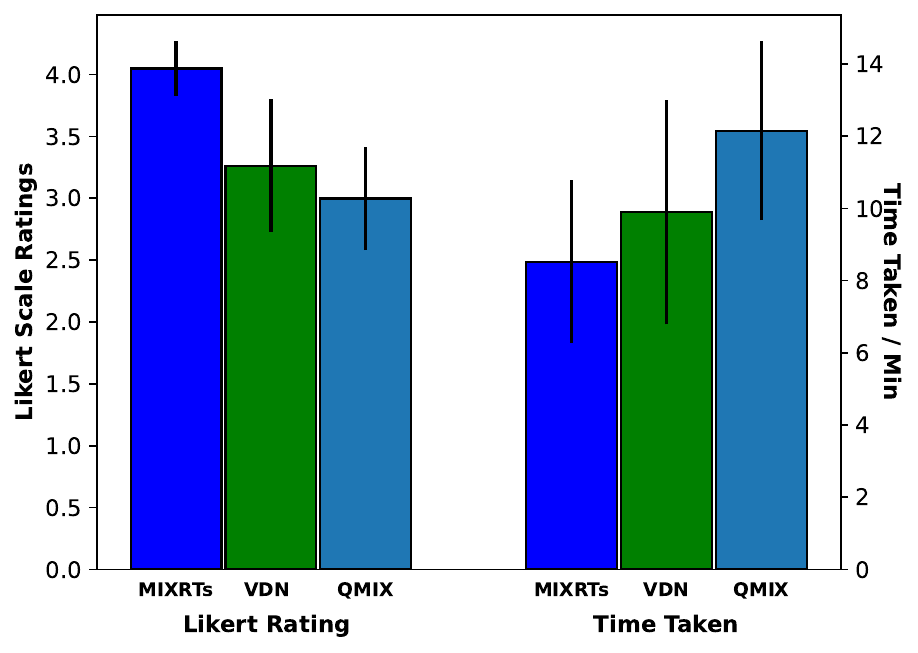}
	\caption{Results of our user study, where a higher Likert rating or lower time taken is better.}
	\label{fucku}
\end{figure}

The results of our user study are shown in Fig.~\ref{fucku}.
Consistent with intuition and previous metrics, human evaluators reveal our MIXRTs a positive helpfulness rating of $4.05$.
The rating indicates that our MIXRTs generally offer a moderate enhancement to interpretability. 
Notably, this result substantially outperforms QMIX's helpfulness rating of $2.9979$ and VDN’s rating of $3.2640$, both of which suggest vague and deviate-related content. 
We conduct an ANOVA on the Likert scale ratings, yielding $F(2, 57) = 33.40, p <
0.0001$. 
This indicates that different mixing methods significantly affect the interpretability and usability of credit assignments.
A Tukey's HSD post-hoc test showed significant differences between MIXRTs and QMIX ($t = 10.26$, $p < 0.0001$) and between MIXRTs and VDN ($t = 7.67$, $p < 0.0001$).
We also measure the time participants took to evaluate each method, as shown in Fig.~\ref{fucku}.
MIXRTs required the least amount of time compared to the other methods.
An interesting observation from the study participants is that some participants expressed they would have abandoned the task if they had known the credits were generated by QMIX. 
These results support the hypothesis that our model significantly outperforms other approaches in reducing human frustration and improving interpretability.
These empirical insights strongly support the supposition that our proposed model markedly excels in diminishing human frustration and enhancing interpretability.

\section{Conclusion}\label{Sec6}
This paper presented MIXRTs, a novel interpretable MARL mixing architecture based on the SDTs with recurrent structure. 
MIXRTs allow end-to-end training in a centralized fashion and learn to linearly factorize a joint action-value function for executing decentralized policies.
The empirical results show that MIXRTs provide not only good model interpretability but also competitive learning performance.
Our approach captures the implicit knowledge of the challenging tasks with a transparent model and facilitates understanding of the learned domain knowledge and how input states influence decisions, which paves the application process in high-stake domains.
Besides, the linear mixing component demonstrates the possibility of generating credit assignments to analyze what role each agent plays among its allies.
Our work motivates future work toward building more interpretable and explainable multi-agent systems.
In future work, we will attempt to analyze higher-level strategies and provide more reliable interpretation estimates, aiming to further reduce the human effort required to understand the decision-making process.

\bibliographystyle{IEEEtran}
\bibliography{reference}

\begin{thebibliography}{10}
\providecommand{\url}[1]{#1}
\csname url@samestyle\endcsname
\providecommand{\newblock}{\relax}
\providecommand{\bibinfo}[2]{#2}
\providecommand{\BIBentrySTDinterwordspacing}{\spaceskip=0pt\relax}
\providecommand{\BIBentryALTinterwordstretchfactor}{4}
\providecommand{\BIBentryALTinterwordspacing}{\spaceskip=\fontdimen2\font plus
\BIBentryALTinterwordstretchfactor\fontdimen3\font minus
  \fontdimen4\font\relax}
\providecommand{\BIBforeignlanguage}[2]{{%
\expandafter\ifx\csname l@#1\endcsname\relax
\typeout{** WARNING: IEEEtran.bst: No hyphenation pattern has been}%
\typeout{** loaded for the language `#1'. Using the pattern for}%
\typeout{** the default language instead.}%
\else
\language=\csname l@#1\endcsname
\fi
#2}}
\providecommand{\BIBdecl}{\relax}
\BIBdecl

\bibitem{vinyals2019grandmaster}
O.~Vinyals, I.~Babuschkin, W.~M. Czarnecki, M.~Mathieu, A.~Dudzik, J.~Chung,
  D.~H. Choi, R.~Powell, T.~Ewalds, P.~Georgiev \emph{et~al.}, ``Grandmaster
  level in {StarCraft II} using multi-agent reinforcement learning,''
  \emph{Nature}, vol. 575, no. 7782, pp. 350--354, 2019.

\bibitem{sunehag2017value}
P.~Sunehag, G.~Lever, A.~Gruslys, W.~M. Czarnecki, V.~Zambaldi, M.~Jaderberg,
  M.~Lanctot, N.~Sonnerat, J.~Z. Leibo, K.~Tuyls \emph{et~al.},
  ``Value-decomposition networks for cooperative multi-agent learning based on
  team reward,'' in \emph{Proceedings of the International Conference on
  Autonomous Agents and MultiAgent Systems}, 2018, pp. 2085--2087.

\bibitem{rashid2018qmix}
T.~Rashid, M.~Samvelyan, C.~S. de~Witt, G.~Farquhar, J.~N. Foerster, and
  S.~Whiteson, ``{QMIX:} {M}onotonic value function factorisation for deep
  multi-agent reinforcement learning,'' in \emph{Proceedings of the
  International Conference on Machine Learning}, vol.~80, 2018, pp. 4295--4304.

\bibitem{li2022metadrive}
Q.~Li, Z.~Peng, L.~Feng, Q.~Zhang, Z.~Xue, and B.~Zhou, ``Metadrive: Composing
  diverse driving scenarios for generalizable reinforcement learning,''
  \emph{IEEE Transactions on Pattern Analysis and Machine Intelligence},
  vol.~45, no.~3, pp. 3461--3475, 2022.

\bibitem{yu2019distributed}
C.~Yu, X.~Wang, X.~Xu, M.~Zhang, H.~Ge, J.~Ren, L.~Sun, B.~Chen, and G.~Tan,
  ``Distributed multiagent coordinated learning for autonomous driving in
  highways based on dynamic coordination graphs,'' \emph{IEEE Transactions on
  Intelligent Transportation Systems}, vol.~21, no.~2, pp. 735--748, 2019.

\bibitem{kiran2021deep}
B.~R. Kiran, I.~Sobh, V.~Talpaert, P.~Mannion, A.~A. Al~Sallab, S.~Yogamani,
  and P.~P{\'e}rez, ``Deep reinforcement learning for autonomous driving: A
  survey,'' \emph{IEEE Transactions on Intelligent Transportation Systems},
  vol.~23, no.~6, pp. 4909--4926, 2021.

\bibitem{liu2021semantic}
I.-J. Liu, Z.~Ren, R.~A. Yeh, and A.~G. Schwing, ``Semantic tracklets: {A}n
  object-centric representation for visual multi-agent reinforcement
  learning,'' in \emph{Proceedings of the IEEE/RSJ International Conference on
  Intelligent Robots and Systems}, 2021, pp. 5603--5610.

\bibitem{10250993}
J.~Wu, Y.~Zhou, H.~Yang, Z.~Huang, and C.~Lv, ``Human-guided reinforcement
  learning with sim-to-real transfer for autonomous navigation,'' \emph{IEEE
  Transactions on Pattern Analysis and Machine Intelligence}, vol.~45, no.~12,
  pp. 14\,745--14\,759, 2023.

\bibitem{topin2021iterative}
N.~Topin, S.~Milani, F.~Fang, and M.~Veloso, ``Iterative bounding {MDP}s:
  {L}earning interpretable policies via non-interpretable methods,'' in
  \emph{Proceedings of the AAAI Conference on Artificial Intelligence},
  vol.~35, no.~11, 2021, pp. 9923--9931.

\bibitem{liu2019tabby}
H.~Liu, R.~Wang, S.~Shan, and X.~Chen, ``What is a tabby? {I}nterpretable model
  decisions by learning attribute-based classification criteria,'' \emph{IEEE
  Transactions on Pattern Analysis and Machine Intelligence}, vol.~43, no.~5,
  pp. 1791--1807, 2019.

\bibitem{zhou2018interpreting}
B.~Zhou, D.~Bau, A.~Oliva, and A.~Torralba, ``Interpreting deep visual
  representations via network dissection,'' \emph{IEEE Transactions on Pattern
  Analysis and Machine Intelligence}, vol.~41, no.~9, pp. 2131--2145, 2018.

\bibitem{hassija2024interpreting}
V.~Hassija, V.~Chamola, A.~Mahapatra, A.~Singal, D.~Goel, K.~Huang,
  S.~Scardapane, I.~Spinelli, M.~Mahmud, and A.~Hussain, ``Interpreting
  black-box models: a review on explainable artificial intelligence,''
  \emph{Cognitive Computation}, vol.~16, no.~1, pp. 45--74, 2024.

\bibitem{tjoa2020survey}
E.~Tjoa and C.~Guan, ``A survey on explainable artificial intelligence ({XAI}):
  {T}oward medical {XAI},'' \emph{IEEE Transactions on Neural Networks and
  Learning Systems}, vol.~32, no.~11, pp. 4793--4813, 2020.

\bibitem{rudin2019stop}
C.~Rudin, ``Stop explaining black box machine learning models for high stakes
  decisions and use interpretable models instead,'' \emph{Nature Machine
  Intelligence}, vol.~1, no.~5, pp. 206--215, 2019.

\bibitem{rudin2022interpretable}
C.~Rudin, C.~Chen, Z.~Chen, H.~Huang, L.~Semenova, and C.~Zhong,
  ``Interpretable machine learning: {F}undamental principles and 10 grand
  challenges,'' \emph{Statistics Surveys}, vol.~16, pp. 1--85, 2022.

\bibitem{xu2022towards}
X.~Xu, Z.~Wang, C.~Deng, H.~Yuan, and S.~Ji, ``Towards improved and
  interpretable deep metric learning via attentive grouping,'' \emph{IEEE
  Transactions on Pattern Analysis and Machine Intelligence}, vol.~45, no.~1,
  pp. 1189--1200, 2022.

\bibitem{cao2019interpretable}
Q.~Cao, X.~Liang, B.~Li, and L.~Lin, ``Interpretable visual question answering
  by reasoning on dependency trees,'' \emph{IEEE Transactions on Pattern
  Analysis and Machine Intelligence}, vol.~43, no.~3, pp. 887--901, 2019.

\bibitem{natarajan2020effects}
M.~Natarajan and M.~Gombolay, ``Effects of anthropomorphism and accountability
  on trust in human robot interaction,'' in \emph{Proceedings of the ACM/IEEE
  International Conference on Human-Robot Interaction}, 2020, pp. 33--42.

\bibitem{puiutta2020explainable}
E.~Puiutta and E.~Veith, ``Explainable reinforcement learning: {A} survey,'' in
  \emph{Proceedings of the International Cross-Domain Conference for Machine
  Learning and Knowledge Extraction}, 2020, pp. 77--95.

\bibitem{shi2020self}
W.~Shi, G.~Huang, S.~Song, Z.~Wang, T.~Lin, and C.~Wu, ``Self-supervised
  discovering of interpretable features for reinforcement learning,''
  \emph{IEEE Transactions on Pattern Analysis and Machine Intelligence},
  vol.~44, no.~5, pp. 2712--2724, 2020.

\bibitem{zheng2024symbolic}
W.~Zheng, S.~Sharan, Z.~Fan, K.~Wang, Y.~Xi, and Z.~Wang, ``Symbolic visual
  reinforcement learning: A scalable framework with object-level abstraction
  and differentiable expression search,'' \emph{IEEE Transactions on Pattern
  Analysis and Machine Intelligence}, 2024.

\bibitem{verma2019verifiable}
A.~Verma, ``Verifiable and interpretable reinforcement learning through program
  synthesis,'' in \emph{Proceedings of the AAAI Conference on Artificial
  Intelligence}, 2019, pp. 9902--9903.

\bibitem{jiang2019neural}
Z.~Jiang and S.~Luo, ``Neural logic reinforcement learning,'' in
  \emph{Proceedings of the International Conference on Machine Learning},
  vol.~97, 2019, pp. 3110--3119.

\bibitem{shi2021temporal}
W.~Shi, G.~Huang, S.~Song, and C.~Wu, ``Temporal-spatial causal interpretations
  for vision-based reinforcement learning,'' \emph{IEEE Transactions on Pattern
  Analysis and Machine Intelligence}, vol.~44, no.~12, pp. 10\,222--10\,235,
  2021.

\bibitem{silver2016mastering}
D.~Silver, A.~Huang, C.~J. Maddison, A.~Guez, L.~Sifre, G.~Van Den~Driessche,
  J.~Schrittwieser, I.~Antonoglou, V.~Panneershelvam, M.~Lanctot \emph{et~al.},
  ``Mastering the game of {G}o with deep neural networks and tree search,''
  \emph{Nature}, vol. 529, no. 7587, pp. 484--489, 2016.

\bibitem{loh2011classification}
W.-Y. Loh, ``Classification and regression trees,'' \emph{Wiley
  Interdisciplinary Reviews: Data Mining and Knowledge Discovery}, vol.~1,
  no.~1, pp. 14--23, 2011.

\bibitem{breiman2001random}
L.~Breiman, ``Random forests,'' \emph{Machine Learning}, vol.~45, no.~1, pp.
  5--32, 2001.

\bibitem{frosst2017distilling}
N.~Frosst and G.~Hinton, ``Distilling a neural network into a soft decision
  tree,'' \emph{arXiv preprint arXiv:1711.09784}, 2017.

\bibitem{silva2020optimization}
A.~Silva, M.~Gombolay, T.~Killian, I.~Jimenez, and S.-H. Son, ``Optimization
  methods for interpretable differentiable decision trees applied to
  reinforcement learning,'' in \emph{Proceedings of the International
  Conference on Artificial Intelligence and Statistics}, vol. 108, 2020, pp.
  1855--1865.

\bibitem{suarez1999globally}
A.~Su{\'a}rez and J.~F. Lutsko, ``Globally optimal fuzzy decision trees for
  classification and regression,'' \emph{IEEE Transactions on Pattern Analysis
  and Machine Intelligence}, vol.~21, no.~12, pp. 1297--1311, 1999.

\bibitem{coppens2019distilling}
Y.~Coppens, K.~Efthymiadis, T.~Lenaerts, A.~Now{\'e}, T.~Miller, R.~Weber, and
  D.~Magazzeni, ``Distilling deep reinforcement learning policies in soft
  decision trees,'' in \emph{Proceedings of the IJCAI Workshop on Explainable
  Artificial Intelligence}, 2019, pp. 1--6.

\bibitem{ding2020cdt}
Z.~Ding, P.~Hernandez-Leal, G.~W. Ding, C.~Li, and R.~Huang, ``{CDT:}
  {C}ascading decision trees for explainable reinforcement learning,''
  \emph{arXiv preprint arXiv:2011.07553}, 2020.

\bibitem{brockman2016openai}
G.~Brockman, V.~Cheung, L.~Pettersson, J.~Schneider, J.~Schulman, J.~Tang, and
  W.~Zaremba, ``Open{AI} {G}ym,'' \emph{arXiv preprint arXiv:1606.01540}, 2016.

\bibitem{samvelyan2019starcraft}
M.~Samvelyan, T.~Rashid, C.~S. De~Witt, G.~Farquhar, N.~Nardelli, T.~G. Rudner,
  C.-M. Hung, P.~H. Torr, J.~Foerster, and S.~Whiteson, ``The starcraft
  multi-agent challenge,'' in \emph{Proceedings of the International Joint
  Conference on Autonomous Agents and MultiAgent Systems}, 2019, pp.
  2186--2188.

\bibitem{wang2020qplex}
J.~Wang, Z.~Ren, T.~Liu, Y.~Yu, and C.~Zhang, ``Q{PLEX}: {D}uplex dueling
  multi-agent {Q}-learning,'' in \emph{Proceedings of the International
  Conference on Learning Representations}, 2021.

\bibitem{oliehoek2016concise}
F.~A. Oliehoek and C.~Amato, \emph{A concise introduction to decentralized
  {POMDP}s}.\hskip 1em plus 0.5em minus 0.4em\relax SpringerBriefs in
  Intelligent Systems. Springer, 2016.

\bibitem{watkins1992q}
C.~J. Watkins and P.~Dayan, ``{Q}-learning,'' \emph{Machine Learning}, vol.~8,
  no.~3, pp. 279--292, 1992.

\bibitem{son2019qtran}
K.~Son, D.~Kim, W.~J. Kang, D.~E. Hostallero, and Y.~Yi, ``{QTRAN:} {L}earning
  to factorize with transformation for cooperative multi-agent reinforcement
  learning,'' in \emph{Proceedings of the International Conference on Machine
  Learning}, vol.~97, 2019, pp. 5887--5896.

\bibitem{liu2023na2q}
Z.~Liu, Y.~Zhu, and C.~Chen, ``{NA$^2$Q}: Neural attention additive model for
  interpretable multi-agent {Q}-learning,'' in \emph{Proceedings of the
  International Conference on Machine Learning}, vol. 202, 2023, pp.
  22\,539--22\,558.

\bibitem{mnih2015human}
V.~Mnih, K.~Kavukcuoglu, D.~Silver, A.~A. Rusu, J.~Veness, M.~G. Bellemare,
  A.~Graves, M.~Riedmiller, A.~K. Fidjeland, G.~Ostrovski \emph{et~al.},
  ``Human-level control through deep reinforcement learning,'' \emph{Nature},
  vol. 518, no. 7540, pp. 529--533, 2015.

\bibitem{irsoy2012soft}
O.~Irsoy, O.~T. Y{\i}ld{\i}z, and E.~Alpayd{\i}n, ``Soft decision trees,'' in
  \emph{Proceedings of the International Conference on Pattern Recognition},
  2012, pp. 1819--1822.

\bibitem{roth2019conservative}
A.~M. Roth, N.~Topin, P.~Jamshidi, and M.~Veloso, ``Conservative
  {Q}-improvement: {R}einforcement learning for an interpretable decision-tree
  policy,'' \emph{arXiv preprint arXiv:1907.01180}, 2019.

\bibitem{pace2021poetree}
A.~Pace, A.~Chan, and M.~van~der Schaar, ``{POETREE}: {I}nterpretable policy
  learning with adaptive decision trees,'' in \emph{Proceedings of the
  International Conference on Learning Representations}, 2021, pp. 1--28.

\bibitem{tan1993multi}
M.~Tan, ``Multi-agent reinforcement learning: {I}ndependent vs cooperative
  agents,'' in \emph{Proceedings of the International Conference on Machine
  Learning}, 1993, pp. 330--337.

\bibitem{oliehoek2008optimal}
F.~A. Oliehoek, M.~T. Spaan, and N.~Vlassis, ``Optimal and approximate
  {Q}-value functions for decentralized {POMDP}s,'' \emph{Journal of Artificial
  Intelligence Research}, vol.~32, pp. 289--353, 2008.

\bibitem{kraemer2016multi}
L.~Kraemer and B.~Banerjee, ``Multi-agent reinforcement learning as a rehearsal
  for decentralized planning,'' \emph{Neurocomputing}, vol. 190, pp. 82--94,
  2016.

\bibitem{hausknecht2015deep}
M.~Hausknecht and P.~Stone, ``Deep recurrent {Q}-learning for partially
  observable {MDP}s,'' in \emph{Proceedings of the AAAI Fall Symposium on
  Sequential Decision Making for Intelligent Agents}, 2015, pp. 29--37.

\bibitem{zhang2012ensemble}
C.~Zhang and Y.~Ma, \emph{Ensemble machine learning: {M}ethods and
  applications}.\hskip 1em plus 0.5em minus 0.4em\relax Springer, 2012.

\bibitem{hazimeh2020tree}
H.~Hazimeh, N.~Ponomareva, P.~Mol, Z.~Tan, and R.~Mazumder, ``The tree ensemble
  layer: {D}ifferentiability meets conditional computation,'' in
  \emph{Proceedings of the International Conference on Machine Learning}, vol.
  119, 2020, pp. 4138--4148.

\bibitem{derbeko2002variance}
P.~Derbeko, R.~El-Yaniv, and R.~Meir, ``Variance optimized bagging,'' in
  \emph{Proceedings of the European Conference on Machine Learning}, 2002, pp.
  60--72.

\bibitem{hasselt2010double}
H.~van Hasselt, ``Double {Q}-learning,'' in \emph{Proceedings of the Advances
  in Neural Information Processing Systems}, vol.~23, 2010, pp. 2613--2621.

\bibitem{van2016deep}
H.~Van~Hasselt, A.~Guez, and D.~Silver, ``Deep reinforcement learning with
  double {Q}-learning,'' in \emph{Proceedings of the AAAI Conference on
  Artificial Intelligence}, vol.~30, no.~1, 2016, pp. 2094--2100.

\bibitem{lowe2017multi}
R.~Lowe, Y.~I. Wu, A.~Tamar, J.~Harb, O.~Pieter~Abbeel, and I.~Mordatch,
  ``Multi-agent actor-critic for mixed cooperative-competitive environments,''
  in \emph{Proceedings of the Advances in Neural Information Processing
  Systems}, vol.~30, 2017.

\bibitem{tampuu2017multiagent}
A.~Tampuu, T.~Matiisen, D.~Kodelja, I.~Kuzovkin, K.~Korjus, J.~Aru, J.~Aru, and
  R.~Vicente, ``Multiagent cooperation and competition with deep reinforcement
  learning,'' \emph{PloS one}, vol.~12, no.~4, pp. 1--15, 2017.

\bibitem{silva2021encoding}
A.~Silva and M.~Gombolay, ``Encoding human domain knowledge to warm start
  reinforcement learning,'' in \emph{Proceedings of the AAAI conference on
  artificial intelligence}, vol.~35, no.~6, 2021, pp. 5042--5050.

\end{thebibliography}

\begin{IEEEbiography}[{\includegraphics[width=1.0in,height=1.25in,clip,keepaspectratio]{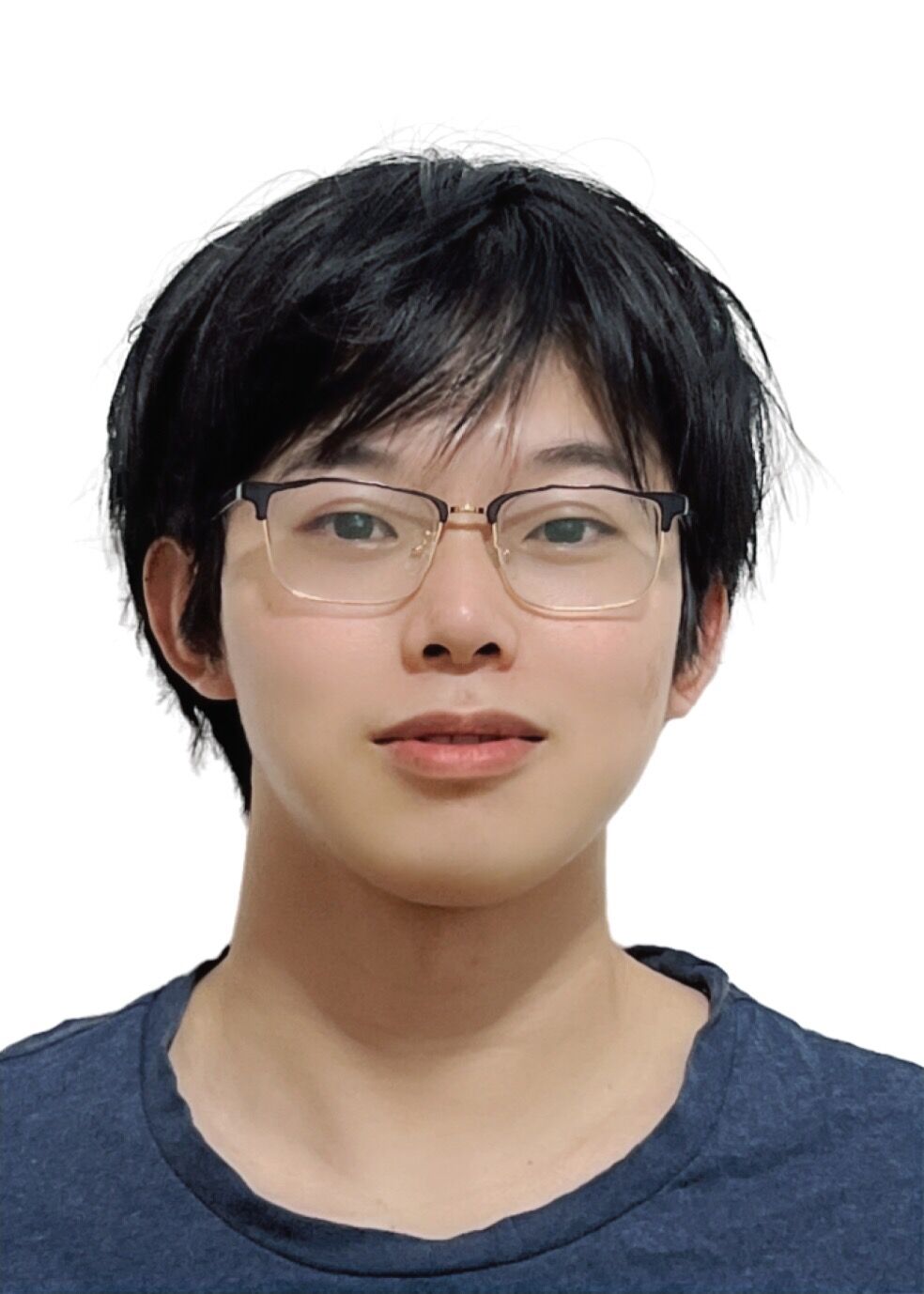}}]{Zichuan Liu} received the B.S. degree in the Computer Science and Technology from Wuhan University of Technology, Wuhan, China, in 2022. 
He is currently pursuing the M.S. degree in the Department of Control Science and Intelligence Engineering, School of Management and Control, Nanjing University, Nanjing, China. 
His research interests include reinforcement learning, data mining, and explainability, with a particular focus on multi-agent reinforcement learning. 
\end{IEEEbiography}

\begin{IEEEbiography}[{\includegraphics[width=1.0in,height=1.25in,clip,keepaspectratio]{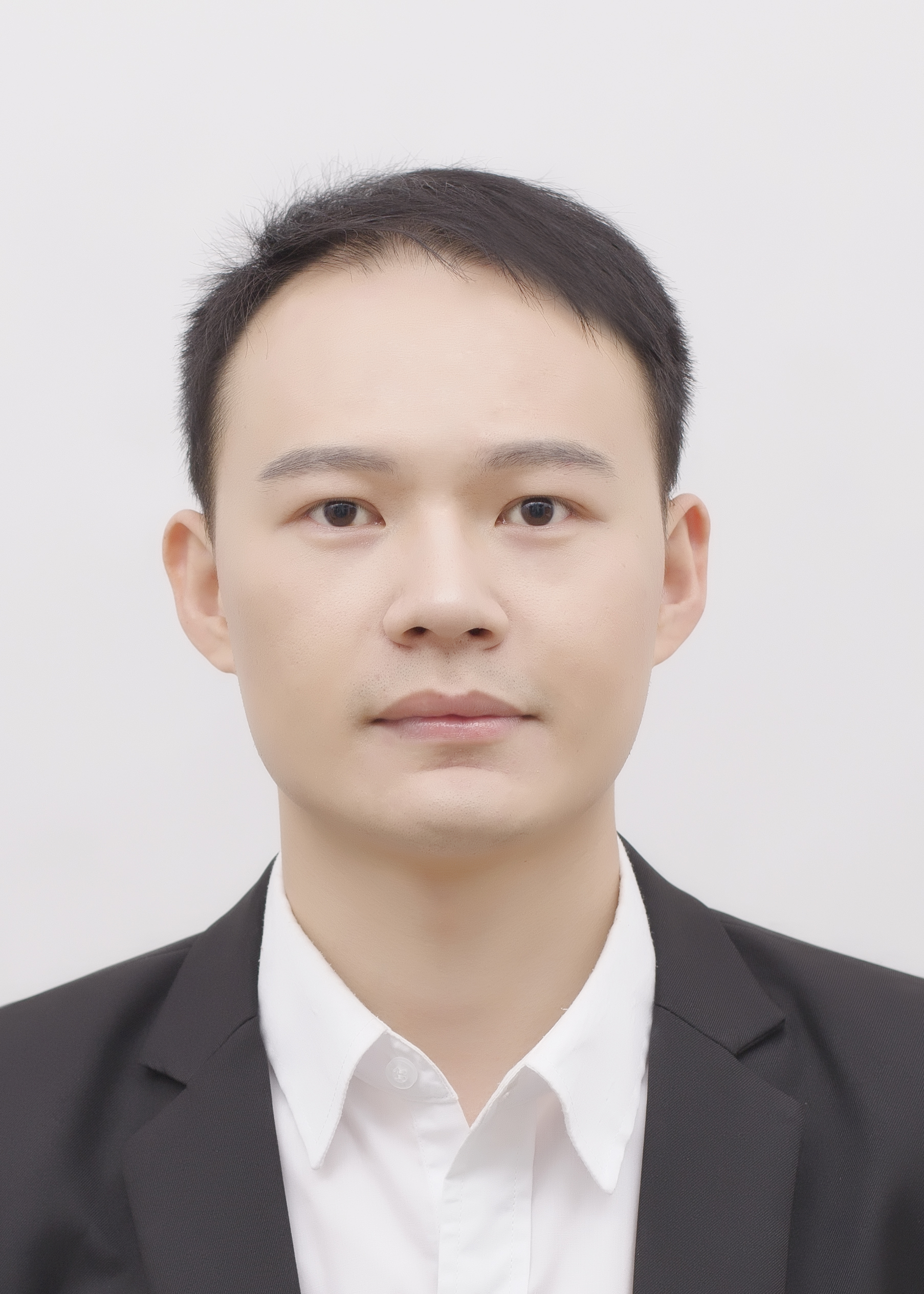}}]{Yuanyang Zhu} (S'19-M'24)
	 received the B.E. degree in automation from the Department of Automation, Huaiyin Institute of Technology, Huai'an, China, in 2017, and the M.S. and Ph.D. degree in the Department of Control Science and Intelligence Engineering, School of Management and Engineering, Nanjing University, Nanjing, China, in 2020 and 2024.
	 He is now a Postdoctoral Fellow at the School of Information Management at Nanjing University.
	 His current research interests include reinforcement learning, machine learning, robotics, and AI for science.
\end{IEEEbiography}

\begin{IEEEbiography}[{\includegraphics[width=1.0in,height=1.25in,clip,keepaspectratio]{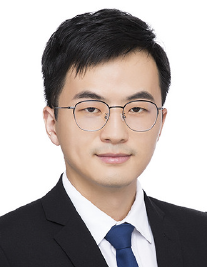}}]{Zhi Wang} (S'19-M'20) received the Ph.D. degree in machine learning from the Department of Systems Engineering and Engineering Management, City University of Hong Kong, Hong Kong, China, in 2019, and the B.E. degree in automation from Nanjing University, Nanjing, China, in 2015. He is currently an Associate Professor with the Department of Control Science and Intelligence Engineering, School of Management and Engineering, Nanjing University, Nanjing, China. He holds visiting positions at the University of New South Wales, Australia and the State Key Laboratory of Management and Control for Complex Systems, Institute of Automation, Chinese Academy of Sciences, China.
 
His current research interests include reinforcement learning, language agents, and embodied intelligence.
He served as the Associate Editor for IEEE International Conference on Systems, Man, and Cybernetics 2023, 2022, and 2021, and the Associate Editor for IEEE International Conference on Networking, Sensing, and Control 2020.
\end{IEEEbiography}
\vfill
\newpage

\begin{IEEEbiography}[{\includegraphics[width=1.0in,height=1.25in,clip,keepaspectratio]{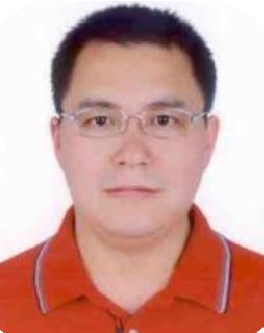}}]{Yang Gao} received the B.S. degree in mechanical
engineering from the Dalian University of Technology, Dalian, China, in 1993, the M.S. degree in computer-aided design from the Nanjing University of Science and Technology, Nanjing, China in 1996, and the Ph.D. degree in computer science from Nanjing University, Nanjing, in 2000, where he is currently a Professor and the Executive Dean of the School of Intelligence Science and Technology.

He is directing the Reasoning and Learning Research Group, Nanjing University. He has published more than 100 articles in top-tired conferences and journals. His research interests include artificial
intelligence and machine learning. He serves as a Program Chair and an Area Chair for many international conferences.
\end{IEEEbiography}

\begin{IEEEbiography}[{\includegraphics[width=1.0in,height=1.25in,clip,keepaspectratio]{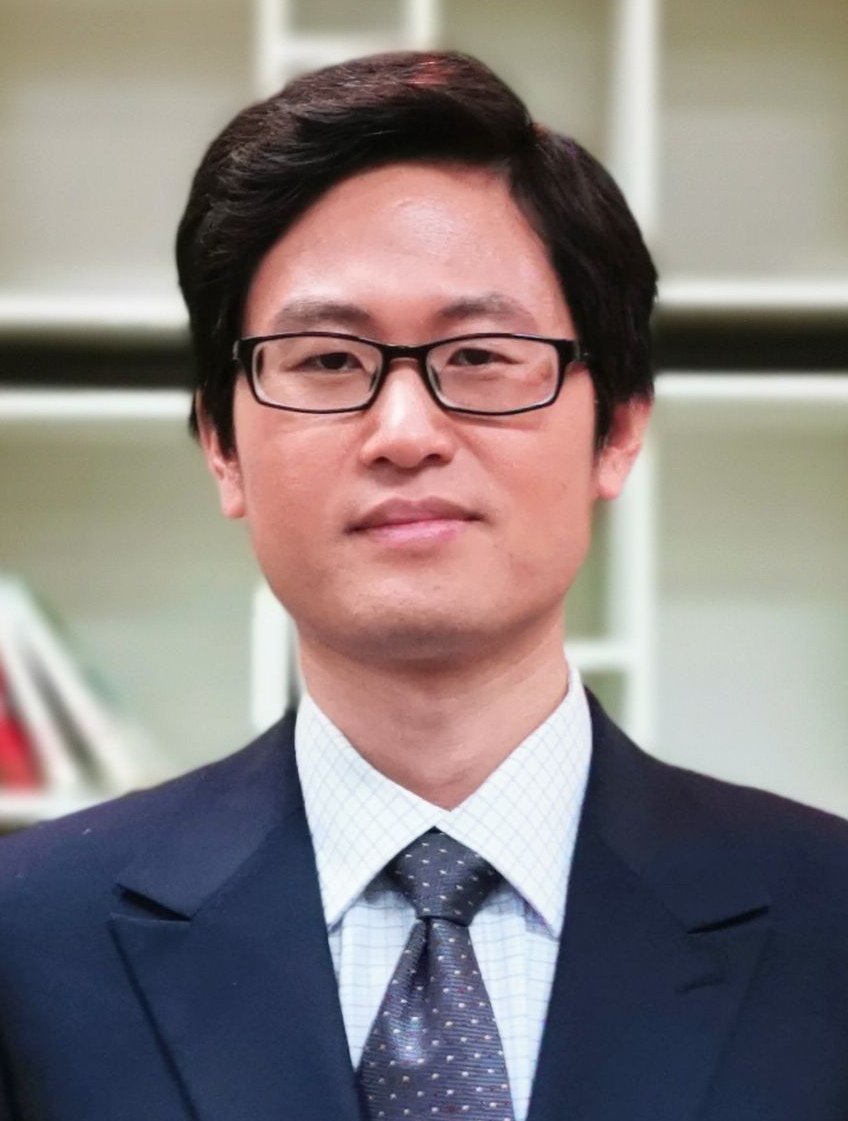}}]{Chunlin Chen}
	(S'05-M'06-SM'21) received the B.E. degree in automatic control and Ph.D. degree in control science and engineering from the University of Science and Technology of China, Hefei, China, in 2001 and 2006, respectively. 
	He is currently a full professor and the vice dean of School of Management and Engineering, Nanjing University, Nanjing, China. 
	He was a visiting scholar at Princeton University, Princeton, USA, from 2012 to 2013. He had visiting positions at the University of New South Wales, Canberra, Australia, and the City University of Hong Kong, Hong Kong, China.
	
	His recent research interests include reinforcement learning, mobile robotics, and quantum control. 
	He is the Chair of Technical Committee on Quantum Cybernetics, IEEE Systems, Man and Cybernetics Society.
\end{IEEEbiography}
\vfill

\end{document}